\RequirePackage{iftex}
\ifPDFTeX\pdfoutput=1\fi
\documentclass[10pt,logo,copyright]{nvidiatechreport}
\usepackage{amsmath,amsfonts,bm}

\def\eqref#1{equation~\ref{#1}}

\def\1{\bm{1}}

\DeclareMathAlphabet{\mathsfit}{\encodingdefault}{\sfdefault}{m}{sl}
\SetMathAlphabet{\mathsfit}{bold}{\encodingdefault}{\sfdefault}{bx}{n}

\usepackage[numbers,sort&compress,square]{natbib}
\usepackage{parskip}
\usepackage{placeins}
\usepackage{etoolbox}
\usepackage{wrapfig}
\usepackage{multirow}
\makeatletter
\newcommand{\FinishWrap}{%
  \par
  \ifnum\c@WF@wrappedlines>1
    \vskip\dimexpr\baselineskip*\numexpr\c@WF@wrappedlines-1\relax\relax
  \fi
  \WFclear
}
\makeatother
\ifPDFTeX
  \AtBeginDocument{\normalfont}
\fi

\patchcmd{\maketitle}{\vskip20pt}{\vskip6pt}{}{%
  \PackageError{grana}{Could not adjust the first-page title spacing}{Check the NVIDIA title definition.}}
\AtEndDocument{%
  \ifnum\getpagerefnumber{fig:fig1-overview}>1\relax
    \PackageError{grana}{The main figure must remain on page 1}{Reduce the opening-page spacing.}%
  \fi
  \ifnum\getpagerefnumber{grana:opening-end}>1\relax
    \PackageError{grana}{The complete abstract must fit below the main figure on page 1}{Reduce the opening-page spacing.}%
  \fi}

\providecommand{\method}{Grana}

\title{Grana: Compositional Reward Modeling for Long-Horizon Robot Manipulation}

\author{%
  \mbox{Zi-ang Cao\textsuperscript{1,2,*}}, \mbox{Kaiyuan Zheng\textsuperscript{1,3,*}},
  \mbox{Yinzhen Xu\textsuperscript{1}}, \mbox{Haonan Duan\textsuperscript{1}},
  \mbox{Jiasheng Gu\textsuperscript{1}}, \mbox{Fengyuan Hu\textsuperscript{1}},
  \mbox{Nan Zhu\textsuperscript{1}}, \mbox{Kun Lei\textsuperscript{1}},
  \mbox{Kevin Lin\textsuperscript{1}}, \mbox{Ruijie Zheng\textsuperscript{1}},
  \mbox{Shenyuan Gao\textsuperscript{1}}, \mbox{Fangqi Zhu\textsuperscript{1}},
  \mbox{Zhe Zhang\textsuperscript{1}}, \mbox{Radha Poovendran\textsuperscript{3}},
  \mbox{Linxi “Jim” Fan\textsuperscript{1,\ensuremath{\dagger}}}, \mbox{Guanzhi Wang\textsuperscript{1,\ensuremath{\dagger}}},
  \mbox{Yuke Zhu\textsuperscript{1,2,\ensuremath{\dagger}}}\\
  {\normalfont\small
    \textsuperscript{1}NVIDIA\quad
    \textsuperscript{2}The University of Texas at Austin\quad
    \textsuperscript{3}University of Washington}\\
  {\normalfont\footnotesize
    \textsuperscript{*}Equal contribution
    \enspace\ensuremath{\cdot}\enspace\textsuperscript{\ensuremath{\dagger}}Equal advising}\\
  {\normalfont\small\url{https://research.nvidia.com/labs/gear/grana/}}%
}

\newdate{granarelease}{21}{10}{2026}
\renewcommand{\today}{\displaydate{granarelease}}
\renewcommand{\copyrightext}{\footerfont
  \ifPDFTeX{\fontfamily{cmr}\selectfont\textcopyright}\else\textcopyright\fi
  \,2026 NVIDIA. All rights reserved.}
\hypersetup{
  pdftitle={Grana: Compositional Reward Modeling for Long-Horizon Robot Manipulation},
  pdfauthor={Zi-ang Cao, Kaiyuan Zheng, Yinzhen Xu, Haonan Duan, Jiasheng Gu, Fengyuan Hu, Nan Zhu, Kun Lei, Kevin Lin, Ruijie Zheng, Shenyuan Gao, Fangqi Zhu, Zhe Zhang, Radha Poovendran, Linxi “Jim” Fan, Guanzhi Wang, Yuke Zhu},
  pdfsubject={Robot reward modeling with structured progress supervision},
  pdfkeywords={Robot manipulation, reward learning, structured progress supervision, long-horizon manipulation}
}

\begin{document}
\frenchspacing
\maketitle

\noindent\begin{minipage}{\linewidth}
  \centering
  \includegraphics[width=0.95\linewidth]{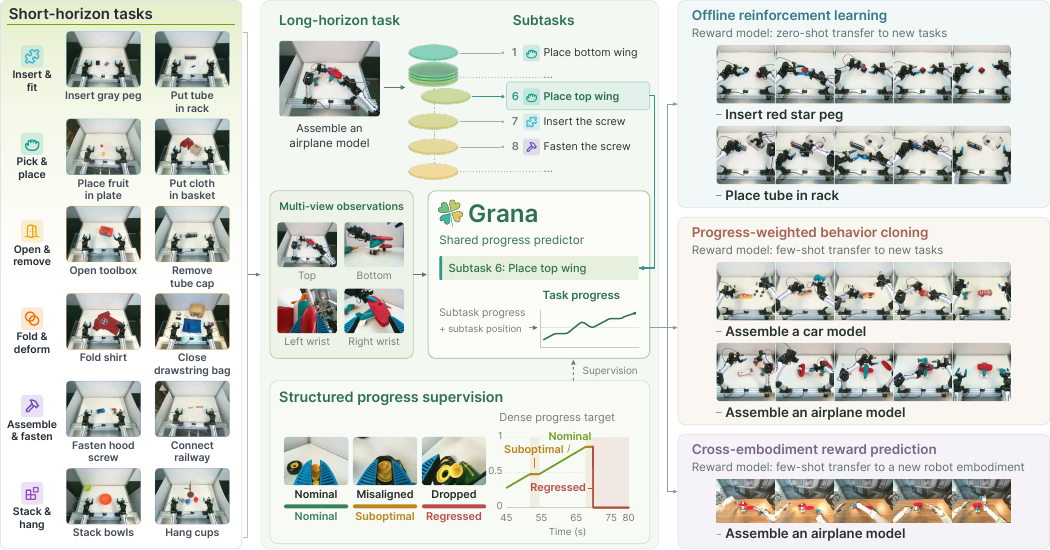}
  \captionsetup{font=footnotesize,skip=3pt,hypcap=false}
  \captionof{figure}{\textbf{Overview of \method{}.}
    Left: diverse short-horizon tasks for pretraining.
    Center: subtask descriptions condition a progress predictor, and sparse annotations define dense progress supervision.
    Each disc in the granum represents a subtask; the offset disc marks the selected subtask.
    Right: offline RL, progress-weighted cloning after long-task adaptation, and cross-embodiment adaptation.}
  \label{fig:fig1-overview}
\end{minipage}\par

\begingroup
  \renewcommand{\absfont}{\bfseries\linespread{1}\fontsize{9}{10.5}\selectfont}
  \begin{abstract}
  Robot reward modeling faces two challenges: generalizing across tasks and accurately tracking progress in long-horizon manipulation.
General-purpose reward models demonstrate broad task transfer, but transfer from short-task pretraining to long-horizon tasks remains underexplored.
Long-horizon progress models often rely on task-specific subtask classifiers or fixed action vocabularies, limiting the reuse of learned progress knowledge when adapting to new tasks.
To address these challenges, we introduce \method{}, a vision-language robot reward model that learns to identify unfinished subtasks and estimate their continuous progress using open-vocabulary subtask descriptions.
We train a language-conditioned model to estimate progress in short-horizon tasks and individual subtasks of long-horizon tasks. For long-horizon tasks, we combine the selected subtask’s predicted progress with credit for preceding completed subtasks to estimate whole-task progress.
General progress prediction requires training the model across diverse tasks with progress supervision that accounts for imperfect execution. Elapsed-time proxies can misrepresent progress when relative subtask durations vary, or failures reverse local progress. We therefore introduce an annotation protocol that converts sparse subtask boundaries and action-quality annotations into dense progress targets, without requiring annotators to assign numerical progress scores.
Pretraining with this supervision on 200 diverse, mostly short-horizon manipulation tasks enables zero-shot reward prediction on unseen tasks and improves progress prediction after adaptation to multi-minute, long-horizon tasks.
Experiments demonstrate (i) generalizable reward prediction, including adaptation to five long-horizon tasks, (ii) benefits from pretraining and composing subtask progress estimates, and (iii) improved downstream policy performance.

  \end{abstract}
  \abscontent\par
  \label{grana:opening-end}
\endgroup
\clearpage

\section{Introduction}
\label{sec:introduction}

Progress-based reward models estimate how much of a task has been completed, providing learning signals for robot policies~\cite{chen2025sarm,liang2026robometer}.
For long-horizon manipulation, these models should reuse progress knowledge learned from shorter tasks while tracking failed attempts and recovery.

Consider assembling a model airplane by placing a wing, inserting a screw, and tightening it with a drill. Completing wing placement should receive the same progress credit across executions, even when the time spent on each subtask varies. During insertion, repeated misalignment should slow progress, while dropping the screw should reset insertion progress without erasing credit for the placed wing. Furthermore, reaching for the drill should not advance progress while insertion remains unfinished. These distinctions motivate us to compose whole-task progress from subtask progress, and to combine subtask boundaries with action-quality annotations to capture setbacks and recovery.

General-purpose reward models learn from diverse robot videos~\cite{alakuijala2025videolanguage,lee2026roboreward,liang2026robometer} or use pretrained vision-language models directly~\cite{chen2026topreward} to provide reward signals across diverse tasks.
We ask whether learning progress at the subtask level improves whole-task progress tracking and supports transfer from short tasks to long-horizon tasks.
Some long-horizon progress models use task-specific subtask classifiers~\cite{chen2025sarm} or a fixed action vocabulary shared across tasks~\cite{chen2026sarm2}.
These choices limit reuse of learned progress knowledge when subtasks change in a new long-horizon task.
We seek a model that learns subtask selection and progress prediction through language descriptions, so that knowledge from short tasks can be reused in longer compositions.

Progress supervision poses a second challenge.
Subtask completion keyframes alone do not reflect failed attempts or recovery.
Local execution judgments can describe setbacks~\cite{mao2026arm,chen2026sarm2}, but a progress curve must also show how much work is lost.
A brief failure may undo the current attempt without undoing earlier subtasks.

We introduce \method{}, a vision-language robot reward model that learns to select the unfinished subtask and predict its local progress from open-vocabulary descriptions (Fig.~\ref{fig:fig1-overview}).
We combine this prediction with credit for preceding completed subtasks to estimate whole-task progress.
Incorrect subtask choices can distort this whole-task estimate, so training considers subtask order to define a local-progress target for each selected subtask.
The same language interface lets us reuse the pretrained model on new tasks and longer compositions without replacing model components.

To train this predictor, we combine subtask boundaries with action-quality annotations to construct a dense progress curve without direct numerical scoring.
Subtask boundaries mark when subtasks are completed, while action-quality annotations describe nominal, suboptimal, and regressed execution within each subtask.
The constructed progress curve advances more slowly during suboptimal execution and drops when regressed execution loses local progress.
It rises again during recovery while preserving credit for earlier completed subtasks.
Using this supervision, we pretrain \method{} on 200 mostly short manipulation tasks.

We evaluate how these design choices affect progress prediction on new tasks and a different embodiment, as well as downstream policy learning.
We also show that structured supervision improves progress predictions from a prior vision-language model (VLM)- based reward model.
We further show \method{} 's generalization capabilities, achieving 20.7\% lower progress RMSE than the strongest evaluated baseline without target-task reward labels or fine-tuning on 50 held-out tasks.
Across five long-horizon tasks with up to 18 subtasks, \method{} adapted with 15 labeled trajectories per task achieves the lowest progress RMSE among the evaluated models.
On the two 3--5-minute airplane and red car assembly tasks, it also outperforms baselines adapted with 100 trajectories; matched comparisons show that pretraining improves prediction on both tasks at both adaptation budgets.
The reward model's zero-shot progress prediction also improves offline reinforcement learning (RL) on held-out tasks.
During iterative policy learning on airplane and red car assembly, progress-weighted behavior cloning improves success over unweighted behavior cloning with the same parent policy and data.

Our contributions are threefold:
\begin{itemize}
    \item \textbf{Compositional progress prediction.} We train the model to select subtasks and predict local progress, then compose their outputs into whole-task progress. During adaptation, task order defines progress targets for the model's own subtask choices.
    \item \textbf{Structured progress supervision.} We combine subtask boundaries and action-quality annotations to construct dense progress curves, with explicit rules for failure and recovery.
    \item \textbf{Evidence for reward model generalization and policy improvement.} We show that short-task pretraining improves long-task adaptation and that predicted progress supports downstream policy learning.
\end{itemize}

\section{Related Work}
\label{sec:related-work}

\paragraph{General-purpose robot reward models}
Pretrained visual and vision-language representations can be adapted to provide reusable reward signals from image goals, text, or demonstration videos~\cite{ma2023vip,ma2023livlanguageimagerepresentationsrewards,rocamonde2024visionlanguage,sontakke2023roboclip}.
Recent approaches use pretrained VLMs to estimate task progress without further fine-tuning~\cite{ma2024generative,chen2026topreward} or learn reward and value predictors from diverse video data~\cite{alakuijala2025videolanguage,lee2026roboreward,liang2026robometer,huang2026rynnvalue}.
However, the benefits of short-task pretraining for long-horizon progress prediction remain underexplored, so we study whether subtask supervision improves long-task adaptation.

\paragraph{Progress supervision}
Video-derived supervision for reward learning includes temporal ordering and transformations~\cite{alakuijala2025videolanguage,zhang2025rewind,yu2026warprm}, signed frame distances or remaining-time targets~\cite{liu2026timerewarder,huang2026rynnvalue}, and successful-transition prediction~\cite{zhang2026dream2reward}.
Other supervision uses subtask boundaries, interval-level progress and completion labels, or simulated failures~\cite{tan2025robodopamine,mao2026arm,fang2026densereward}.
To capture local setbacks and recovery while preserving completed work, we deterministically convert sparse subtask boundaries and action-quality judgments into dense progress curves without direct numerical scoring.
The constructed curves reset local progress on failure and rebuild it according to action quality during retries, while retaining credit for completed subtasks.

\paragraph{Progress estimation through task decomposition}
REDS infers subtasks through video--language alignment and learns subtask-conditioned rewards from segmented demonstrations~\cite{kim2025reds}.
SARM reports that REDS's predictions remain largely step-shaped despite monotonicity regularization, limiting fine-grained progress estimation. By separately predicting subtask identity and continuous local progress, SARM achieves lower progress-estimation error on T-shirt folding~\cite{chen2025sarm}.
SARM2 extends subtask-aware reward modeling to multiple tasks~\cite{chen2026sarm2}.
PRM4RL composes subtask indices and local progress through generated simulator-state functions, while RoboMeter advances through VLM-proposed subtasks using success predictions during online RL~\cite{zhang2025prm4rl,liang2026robometer}.
\method{} studies how short-task pretraining supports longer compositions by combining predicted local progress with credit for preceding completed subtasks.
Its supervision explicitly captures local failure resets and recovery, while task order defines progress targets even for incorrect subtasks.

\paragraph{Reward-guided policy learning}
Learned visual rewards already support robot policy improvement through RL~\cite{zhang2025rewind,liang2026robometer,zhai2025vlac}.
Offline RL and advantage-weighted learning provide complementary policy-improvement tools~\cite{kostrikov2022offline,zhuang2023behavior,peng2019advantage}.
We evaluate zero-shot reward models for offline RL on short tasks and adapted progress predictors for progress-weighted behavior learning on long-horizon tasks.

\section{Method}
\label{sec:method}

\method{} learns to select the unfinished subtask and predict its local progress, then combines these outputs into whole-task progress.
The same language-conditioned model is pretrained on mostly short tasks and reused for longer compositions.
Its progress targets account for both execution quality and the evaluated subtask's position in the task sequence.

\begin{figure}[!htb]
  \centering
  \captionsetup{font=small,skip=4pt}
  \includegraphics[width=\textwidth]{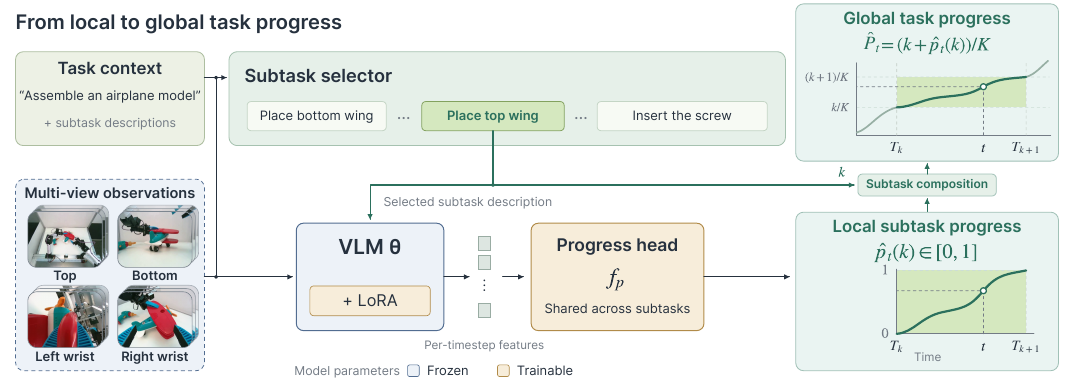}
  \caption{\textbf{From local to global task progress in \method{}.}
  \method{} learns subtask selection and local-progress prediction with a shared vision-language model.
  It selects the unfinished subtask from language descriptions, estimates progress for the selected description, and combines this estimate with credit for completed subtasks.}
  \label{fig:rhoval-language-interface}
\end{figure}

\subsection{Predicting Progress through Subtasks}
\label{sec:problem_formulation}
\label{sec:local_reward_model}
\label{sec:atomic_grounding}
\label{sec:stage_composition}

Given camera observations, a task description, and the task's subtask sequence, \method{} selects the unfinished subtask and predicts its local progress (Fig.~\ref{fig:rhoval-language-interface}).
The supplied sequence specifies the user's preferred execution order and evaluation granularity.
Subtask descriptions are language inputs to a shared predictor; their order and number are not hard-coded into the architecture.
The same task can therefore be queried with different valid orders or coarser or finer decompositions.
Training or adaptation uses supervision defined under the chosen specification, while inference conditions on the supplied list.
Whole-task progress measures completion relative to this user-specified procedure.

Let $k_t$ be the selected subtask's zero-based index, and let $\hat p_t(k_t)\in[0,1]$ be its predicted local subtask progress.
For a task with $K$ subtasks, we assign each subtask an equal share of global whole-task progress:
\begin{equation}
  \hat P_t=\frac{k_t+\hat p_t(k_t)}{K}.
  \label{eq:composed-progress}
\end{equation}
Six completed subtasks and halfway progress through the seventh give $6.5/9$ in a nine-subtask task.
If local progress falls to zero, the estimate becomes $6/9$.
This composition assumes that preceding subtasks remain completed after a local failure; an incorrect selection can still distort the estimate.

\paragraph{Shared predictor}
\label{sec:implementation-details}
In our model, subtask selection and progress prediction share a VLM that processes multiview observations sampled over time, providing context to distinguish a failed attempt from its recovery.
A temporal transformer combines features for local-progress prediction through a shared head.
Temporal aggregation can use later observations. In our policy-learning experiments, we score recorded trajectories offline.

\Needspace{370pt}
\subsection{Supervision for Imperfect Execution}
\label{sec:structured_supervision}
\label{sec:target-construction}

\begingroup
\setlength{\intextsep}{3pt}
\setlength{\columnsep}{12pt}
\begin{wrapfigure}{r}{0.49\textwidth}
  \centering
  \includegraphics[width=\linewidth]{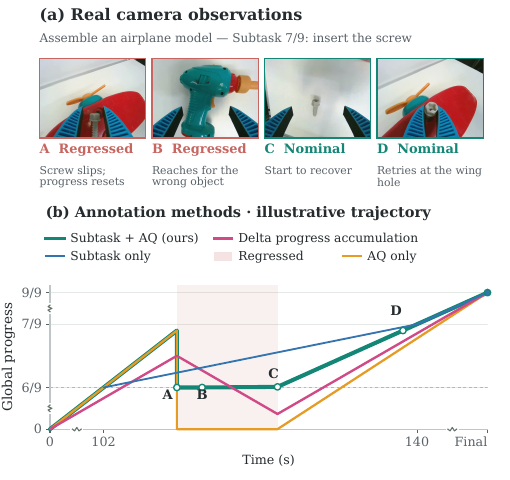}
  \captionsetup{font=footnotesize,skip=3pt}
  \caption{\textbf{Why combine subtask completion and action quality (AQ)?}
  Six of nine subtasks are complete before the screw-insertion failure at A--B and recovery at C--D (top).
  The illustrative curves (bottom) compare progress reconstruction from sparse annotations.
  Subtask only misses the failure, while AQ only drops to zero and discards completed-subtask progress.
  Delta progress accumulates positive, zero, or negative changes, following prior work~\cite{mao2026arm}.
  The plotted curve falls gradually after the screw slips, leaving progress near its pre-failure value; without subtask completion anchors, prolonged regression could also drive it below $6/9$.
  Subtask + AQ (ours) immediately resets insertion progress while retaining the $6/9$ from completed subtasks, then tracks recovery.}
  \label{fig:annotation-retry}
\end{wrapfigure}

Selecting the right subtask does not reveal how much of it is complete.
Elapsed time advances through both a successful insertion and repeated misalignment, yet these executions should receive different progress targets.
We therefore combine subtask annotations with judgments of local action quality (AQ).
Subtask annotations identify completed work.
AQ judgments describe setbacks and recovery within the unfinished subtask.

Annotators mark subtask boundaries and intervals of \emph{suboptimal} or \emph{regressed} execution; unmarked intervals are \emph{nominal}.
Nominal execution follows the task normally.
Suboptimal execution is degraded without physical failure, such as a misaligned screw that remains held.
Regressed execution loses local progress through physical failure, such as dropping the screw.
Separately, each subtask receives a binary outcome: 1 for successful completion and 0 for failure.
These outcomes describe how the subtask ends: a completed subtask can still contain regressed intervals followed by recovery.

The annotations define a local progress target $p_t\in[0,1]$ without requiring numerical scores for individual frames.
Within each uninterrupted attempt, we distribute progress according to action quality, giving nominal frames larger increments than suboptimal frames.
In a successful subtask with retries, regression resets local progress to zero, and a subsequent attempt rebuilds it.
The binary outcome anchors the final local-progress target.

Let $k_t^\star$ denote the annotated subtask to evaluate.
We construct the whole-task progress target $P_t^\star=(k_t^\star+p_t)/K$ using the same composition as at inference.
In the insertion retry in Fig.~\ref{fig:annotation-retry}, the six completed subtasks retain their credit while insertion fails and recovers.

\FinishWrap
\endgroup

\subsection{Training Subtask Selection and Progress Prediction}
\label{sec:reward_model_inference}
\label{sec:progress-training-details}
\label{sec:native-stage-training}

Because \method{} selects its own subtask, progress training must account for a description that is chosen incorrectly.
For example, if insertion is only partly complete but the model selects fastening, fastening should receive zero progress rather than insertion's partial progress.
We use task order to define the target for each queried subtask: earlier completed subtasks receive one, the annotated unfinished subtask receives its local progress, and later subtasks receive zero:
\begin{equation}
  y_t(k_t)=
  \begin{cases}
    1, & k_t<k_t^\star,\\
    p_t, & k_t=k_t^\star,\\
    0, & k_t>k_t^\star.
  \end{cases}
  \label{eq:focus-progress-target}
\end{equation}

Pretraining combines subtask selection with annotated-subtask and whole-task progress prediction.
Selection learns to identify $k_t^\star$, while local-progress training teaches the predictor to evaluate its description.
Whole-task prediction provides auxiliary supervision from available overall labels, allowing data without reliable subtask annotations to contribute.
During adaptation, we also query the progress predictor with descriptions selected by the model itself.
For each selected description, Eq.~\ref{eq:focus-progress-target} provides the corresponding local-progress target, including when the selected subtask is earlier or later than the annotated one.

We implement the shared predictor with Qwen3.5-4B~\cite{qwen2026qwen35} and LoRA~\cite{hu2022lora}, training the adapters, temporal transformer, and progress head while keeping the base VLM frozen.
Selection and progress use cross-entropy objectives.
For progress training, scalar targets are encoded as distributions over neighboring bins; the head predicts a distribution whose mean gives a continuous progress estimate.
Both objectives place greater weight on predictions farther from the target.
Pretraining and adaptation retain the same architecture and language interface.
At inference, the model receives observations, the description, and the subtask list, without annotation inputs.

\subsection{Using Progress for Policy Learning}
\label{sec:policy-learning-method}

Progress differences provide feedback about recorded actions: an increase indicates advancement, while a decrease indicates a setback.
We use these differences in two ways, holding the conversion procedure fixed across reward models in each comparison.

\paragraph{Offline reinforcement learning}
Using the learned progress model $\hat P(s)$, we assign each
$H$-step action chunk $a_{t:t+H}$ a reward equal to the
discounted sum of predicted progress increments:
\[
r_t(s_t,a_{t:t+H})
=
\sum_{j=0}^{H-1}\gamma^j
\left[
\hat P(s_{t+j+1})-\hat P(s_{t+j})
\right],
\]
where $\gamma$ is the per-step discount factor.
Using these rewards, IQL~\cite{kostrikov2022offline} learns an action-value function $Q_{\phi}(s,a_{t:t+H})$ and a state-value function $V_{\psi}(s)$.
We use the resulting advantage estimates, $A(s,a_{t:t+H})=Q_{\phi}(s,a_{t:t+H})-V_{\psi}(s)$, to update the policy with BPPO~\cite{zhuang2023behavior}.

\paragraph{Progress-weighted behavior cloning}
Even successful trajectories can contain unhelpful actions, so we use progress changes to weight their contribution to imitation learning.
We map these progress changes, $\hat P(s_{t+1}) - \hat P(s_{t})$, over each action step to high, intermediate, or low nonnegative imitation weights, using separate thresholds for human and policy actions.
The weighted action loss is normalized by the total weight.
Appendix~\ref{app:target-construction} gives the complete target-construction rules and a worked failure-and-recovery example.
Appendices~\ref{app:model-training}--\ref{app:weighted-bc} specify the model, training settings, and policy-learning conversions.

\FloatBarrier
\section{Experiments}
\label{sec:experiments}

Our experiments address four questions:
\begin{enumerate}
    \item \textbf{Supervision:} Can structured subtask and quality annotations provide practical and useful progress supervision?
    \item \textbf{Generalization:} How well does \method{} generalize across tasks, horizons, embodiments, and datasets?
    \item \textbf{Pretraining and composition:} How does reward-model pretraining improve long-horizon adaptation, and does subtask composition improve progress prediction in both \method{} and a frontier VLM such as GPT-6 Astra~\cite{openai2026gpt6}?
    \item \textbf{Policy learning:} Do the resulting progress estimates improve downstream policy learning?
\end{enumerate}
We first test the supervision used to train the models (Sec.~\ref{sec:annotation-supervision}), then evaluate zero-shot prediction, adaptation, and predictor reuse (Sec.~\ref{sec:reward-prediction}), examine subtask composition (Sec.~\ref{sec:pretraining-composition}), and test policy learning (Sec.~\ref{sec:policy-improvement}).

\subsection{Experimental Setup}
\label{sec:evaluation-protocol}

\begin{figure}[!htb]
  \centering
  \includegraphics[width=\linewidth]{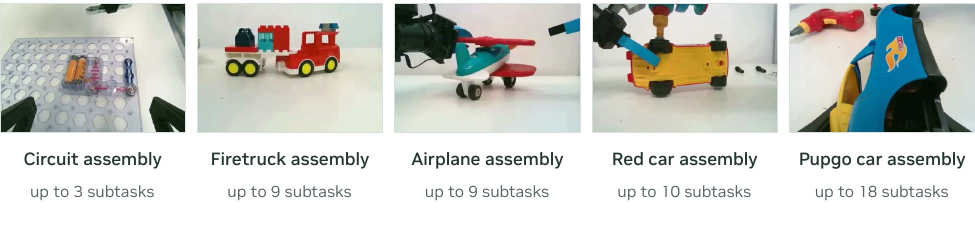}
  \captionsetup{font=footnotesize,skip=5pt}
  \caption{\textbf{Five tasks for long-horizon reward prediction.}
    Selected camera views illustrate the task objects; labels give the maximum number of subtasks.
    Table~\ref{tab:five-task-rmse} reports progress-prediction error on unseen test trajectories for all five tasks.}
  \label{fig:five-long-horizon-tasks}
\end{figure}

Zero-shot evaluation uses 50 held-out tasks.
Long-horizon reward prediction covers five tasks: circuit assembly and switch activation, firetruck assembly, airplane assembly, red car assembly, and Pupgo car assembly (Fig.~\ref{fig:five-long-horizon-tasks}).
We evaluate models adapted with 15 labeled trajectories per task on 24--30 unseen test trajectories per task.
The 3--5-minute airplane ($K=9$) and red car ($K=10$) tasks also support experiments on larger adaptation budgets, pretrained initialization, and policy learning.
Throughout the remaining experiments, ``car model assembly'' refers to the red car task.

We use RMSE to measure errors in predicted progress against constructed curves that preserve completed work and capture failures and recovery.
Across ours and 3 other baseline reward models, lower aggregate RMSE on the 50 held-out tasks corresponds to higher or tied success on both downstream tasks (Sec.~\ref{sec:short-task-policy}), supporting its use to prioritize models for policy evaluation.
Long-task errors use subtask units ($K$ times normalized RMSE on $[0,1]$).

Policy experiments use the GR00T-2 world-action model~\cite{nvidia2026gr00t2}, holding initialization, data, and learning procedure fixed within each task and comparison while varying reward models.

\Needspace{210pt}
\subsection{Annotation Quality and Supervision Utility}
\label{sec:annotation-supervision}

We first measure annotation effort, then examine the resulting target curves, and finally test whether these targets improve reward-model learning.

\begin{table}[!htb]
  \centering
  \begin{minipage}[t]{0.43\linewidth}
    \vspace{0pt}
    \captionsetup{font=footnotesize,skip=5pt}
    \caption{Annotation acceptance and time (minutes per video minute; mean $\pm$ SD).}
    \label{tab:annotation-interface-results}
  \end{minipage}\hfill
  \begin{minipage}[t]{0.53\linewidth}
    \vspace{0pt}
    \captionsetup{font=footnotesize,skip=5pt}
    \caption{RoboMeter progress consistency under different supervision. Metrics are defined in Appendix~\ref{app:progress-consistency-metrics}.}
    \label{tab:progress-consistency}
  \end{minipage}
  \par\noindent
  \begin{minipage}[t]{0.43\linewidth}
    \vspace{0pt}
    \footnotesize
    \setlength{\tabcolsep}{3pt}
    \renewcommand{\arraystretch}{1.12}
    \begin{tabular*}{\linewidth}{@{\extracolsep{\fill}}lrr@{}}
      \toprule
      \textbf{Annotation} & \textbf{Accept.}$\uparrow$ & \textbf{Time}$\downarrow$ \\
      \midrule
      Direct points & 0.76 & $2.6 \pm 0.6$ \\
      AQ only & 0.90 & $1.4 \pm 0.3$ \\
      Subtask only & 0.96 & $1.1 \pm 0.2$ \\
      Subtask + AQ (ours) & 0.90 & $1.5 \pm 0.4$ \\
      \bottomrule
    \end{tabular*}
    \end{minipage}\hfill
  \begin{minipage}[t]{0.53\linewidth}
    \vspace{0pt}
    \footnotesize
    \setlength{\tabcolsep}{2pt}
    \renewcommand{\arraystretch}{1.12}
    \begin{tabular*}{\linewidth}{@{\extracolsep{\fill}}lrrrr@{}}
      \toprule
      & \multicolumn{2}{c}{\textbf{Airplane}} & \multicolumn{2}{c}{\textbf{Car}} \\
      \cmidrule(lr){2-3}\cmidrule(lr){4-5}
      \textbf{Supervision} & Q-Rank$\uparrow$ & D--U$\uparrow$ & Q-Rank$\uparrow$ & D--U$\uparrow$ \\
      \midrule
      Official~\cite{liang2026robometer} & 69.0 & 57.9 & 44.5 & 68.5 \\
      Original target & 63.9 & 59.7 & 76.0 & 49.3 \\
      Our target & \textbf{83.8} & \textbf{70.2} & \textbf{87.0} & \textbf{84.9} \\
      \bottomrule
    \end{tabular*}
    \end{minipage}
\end{table}

\subsubsection{Annotation Interfaces}
\label{sec:annotation-interface-study}

To measure annotation effort and protocol acceptance, we compare numerical scoring (Direct points), subtask boundaries (Subtask only), action-quality windows (AQ only), and their combination (Subtask + AQ).
Quality windows mark suboptimal or regressed execution; unmarked windows are nominal, as defined in Sec.~\ref{sec:structured_supervision}.
The study uses 100 trajectories and 12 annotators, each producing one annotation per trajectory, with each annotation jointly assessed by two additional reviewers.
Acceptance is the fraction that follows the predefined protocol without revision.
Subtask + AQ raises acceptance from 0.76 to 0.90 and reduces mean annotation time from 2.6 to 1.5 minutes per video minute (Table~\ref{tab:annotation-interface-results}).

Subtask only is faster and has higher acceptance, but boundaries alone do not describe failures within an unfinished subtask.
On clean successful trajectories without suboptimal or regressed segments, it can match Subtask + AQ given the same boundaries.

We next compare progress reconstruction from sparse annotations using the accumulation rule used in prior work~\cite{mao2026arm}, subtask or action-quality annotations alone, and their combination (Fig.~\ref{fig:annotation-retry}).
Together with Table~\ref{tab:annotation-interface-results}, this comparison motivates Subtask + AQ: it captures sudden losses of local progress while preserving completed subtasks, and achieves higher efficiency and protocol acceptance than direct numerical scoring.

\subsubsection{Progress Consistency under Different Supervision}
\label{sec:target-intervention}

We evaluate the official RoboMeter checkpoint and RoboMeter trained with its original or our constructed targets using two metrics (Appendix~\ref{app:progress-consistency-metrics}).
Q-Rank measures how often a high-quality keyframe scores above a low-quality keyframe of the same subtask from another episode; a random ordering of each pair is correct half the time on average (50\% accuracy).
Down--Up measures how often scores around an annotated failure satisfy the drop-and-recovery criterion in the appendix.
We report both because random scores can attain high Down--Up scores without consistently ranking execution quality, whereas scores that never decrease always give Down--Up $=0$.
Constructed-target finetuning achieves the highest scores on both tasks (Table~\ref{tab:progress-consistency}).

\FloatBarrier
\subsection{Reward Prediction and Generalization}
\label{sec:reward-prediction}

We evaluate zero-shot prediction on held-out tasks, adaptation to longer tasks and a different embodiment, and whole-trajectory ranking on a public benchmark.

\Needspace{245pt}
\subsubsection{Zero-Shot Prediction and Pretraining Coverage}
\label{sec:pretrain-checkpoints}
\label{sec:pretraining-budget-ablation}

On the 50 held-out tasks without target-task reward labels or fine-tuning, \method{} achieves RMSE 0.180 versus 0.227 for the strongest baseline, RoboMeter, a 20.7\% reduction (Fig.~\ref{fig:scaling}(a)).
Increasing pretraining from 15 to 200 tasks reduces held-out RMSE by 35.3\% (Fig.~\ref{fig:scaling}(b)).

\par\vspace{6pt}\noindent\begin{minipage}{\linewidth}
  \begin{minipage}[t]{0.65\linewidth}
    \vspace{0pt}\centering
    \includegraphics[width=\linewidth]{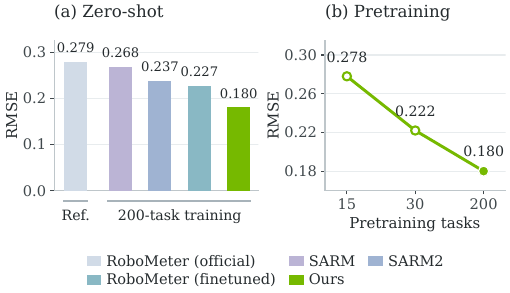}
  \end{minipage}\hfill
  \begin{minipage}[t]{0.32\linewidth}
    \vspace{0pt}
    \captionsetup{font=footnotesize,skip=0pt,hypcap=false}
    \captionof{figure}{Zero-shot progress prediction and pretraining scale. Both panels report RMSE on the same 50 held-out tasks without target-task reward labels or fine-tuning.
      (a) Models trained on the shared 200-task corpus, with the released RoboMeter checkpoint (official) shown as an external reference.
      RoboMeter (finetuned) continues training from this checkpoint.
      (b) \method{} trained on nested sets of 15, 30, and 200 tasks, with training compute and other settings fixed.
      Task budgets are displayed as equally spaced conditions.
      Lower RMSE is better; the panels use different y-axis scales.}
    \label{fig:scaling}
  \end{minipage}
\end{minipage}\par

\Needspace{260pt}
\subsubsection{Long-Horizon Adaptation}
\label{sec:long-horizon-reward-modeling}

We first compare progress prediction across all five long-horizon tasks, adapting \method{}, SARM, SARM2, and RoboMeter with 15 labeled target-task trajectories each (D15).
We also include GPT-6 with subtask composition as a reference.
\method{} achieves the lowest RMSE on every task, reducing error by 16.1--31.6\% relative to the lowest-error comparison model for each task (Table~\ref{tab:five-task-rmse}).

\begin{table}[!htb]
  \centering
  \captionsetup{font=small,skip=5pt}
  \caption{Progress prediction on five long-horizon tasks.
    D15 denotes 15 labeled target-task adaptation trajectories for the learned reward models.
    Evaluation uses 24--30 unseen test trajectories per task.
    RMSE is in subtask units; lower is better.}
  \label{tab:five-task-rmse}
  \footnotesize
  \setlength{\tabcolsep}{3pt}
  \renewcommand{\arraystretch}{1.18}
  \begin{tabular*}{\linewidth}{@{\extracolsep{\fill}}lrrrrr@{}}
    \toprule
    \textbf{Method}
      & \shortstack{\textbf{Circuit}\\\textbf{assembly}}
      & \shortstack{\textbf{Firetruck}\\\textbf{assembly}}
      & \shortstack{\textbf{Airplane}\\\textbf{assembly}}
      & \shortstack{\textbf{Red car}\\\textbf{assembly}}
      & \shortstack{\textbf{Pupgo car}\\\textbf{assembly}} \\
    Subtasks (up to) & 3 & 9 & 9 & 10 & 18 \\
    \midrule
    GPT-6: composition & 0.2920 & 0.2914 & 0.4505 & 0.5240 & 1.8135 \\
    \midrule
    SARM (D15) & 0.3612 & 0.5406 & 0.7474 & 1.1240 & 0.4042 \\
    SARM2 (D15) & 0.2975 & 0.5166 & 0.4980 & 0.6290 & 0.3885 \\
    RoboMeter (D15) & 0.1864 & 0.5202 & 0.4930 & 0.6580 & 0.3189 \\
    Ours (D15) & \textbf{0.1563} & \textbf{0.2264} & \textbf{0.3080} & \textbf{0.3940} & \textbf{0.2561} \\
    \bottomrule
  \end{tabular*}
\end{table}

\FloatBarrier
\Needspace{220pt}
\begingroup
\setlength{\intextsep}{3pt}
\setlength{\columnsep}{12pt}
\begin{wraptable}{r}{0.54\textwidth}
  \centering
  \captionsetup{font=footnotesize,skip=5pt}
  \caption{Adaptation budgets on airplane and red car assembly. RMSE is in subtask units; D15/D100 use 15/100 labeled trajectories.}
  \label{tab:long-horizon-rmse}
  \label{tab:ours-forward-transfer}
  \footnotesize
  \setlength{\tabcolsep}{3pt}
  \renewcommand{\arraystretch}{1.12}
  \begin{tabular*}{\linewidth}{@{\extracolsep{\fill}}lrrrr@{}}
    \toprule
    & \multicolumn{2}{c}{\textbf{Airplane ($K=9$)}} & \multicolumn{2}{c}{\textbf{Car ($K=10$)}} \\
    \cmidrule(lr){2-3}\cmidrule(lr){4-5}
    \textbf{Method} & \textbf{D15}$\downarrow$ & \textbf{D100}$\downarrow$ & \textbf{D15}$\downarrow$ & \textbf{D100}$\downarrow$ \\
    \midrule
    RoboMeter & 0.493 & 0.487 & 0.658 & 0.549 \\
    SARM & 0.747 & 0.485 & 1.124 & 0.539 \\
    SARM2 & 0.498 & 0.461 & 0.629 & 0.472 \\
    \midrule
    Ours: no-pretrain & 0.382 & 0.311 & 0.502 & 0.445 \\
    Ours & \textbf{0.308} & \textbf{0.262} & \textbf{0.394} & \textbf{0.319} \\
    \bottomrule
  \end{tabular*}
\end{wraptable}

To examine data efficiency, we compare the learned reward models on airplane and red car assembly using 15 (D15) or 100 (D100) training trajectories per task, sharing annotations and constructed global progress targets.
\method{} achieves the lowest RMSE in all four task--budget settings and, with D15, outperforms every baseline using D100 (Table~\ref{tab:long-horizon-rmse}).
Averaged in subtask units across these two tasks, RMSE is 37.7\% below SARM2, the strongest adapted baseline in this comparison, at both budgets.
Section~\ref{sec:forward-transfer} analyzes pretrained initialization.

\FinishWrap
\endgroup

\Needspace{150pt}
\subsubsection{Cross-Embodiment Adaptation}
\label{sec:cross-embodiment-transfer}

\begingroup
\setlength{\intextsep}{0pt}
\setlength{\columnsep}{12pt}
\begin{wraptable}{r}{0.48\textwidth}
  \centering
  \captionsetup{font=footnotesize,skip=3pt}
  \caption{YAM to Tianji adaptation: 10 labeled trajectories (D10); RMSE on the $0$--$9$ scale.}
  \label{tab:tianji-transfer}
  \footnotesize
  \setlength{\tabcolsep}{3pt}
  \renewcommand{\arraystretch}{1.12}
  \begin{tabular*}{\linewidth}{@{\extracolsep{\fill}}lrrrr@{}}
    \toprule
    & \textbf{SARM} & \textbf{SARM2} & \textbf{RoboMeter} & \textbf{Ours} \\
    \midrule
    \textbf{RMSE}$\downarrow$ & 0.919 & 1.167 & 0.556 & \textbf{0.452} \\
    \bottomrule
  \end{tabular*}
\end{wraptable}

We transfer reward models fine-tuned for airplane model assembly on the YAM robot to the same task on the Tianji robot, using 10 Tianji trajectories, and evaluate them on the same 30 Tianji test trajectories.
\method{} reaches RMSE 0.452 versus 0.556 for the strongest baseline, RoboMeter, an 18.7\% reduction (Table~\ref{tab:tianji-transfer}).

\FinishWrap
\endgroup

\subsubsection{Reusing Pretrained Knowledge through Language}
\label{sec:transfer-recipes}
\label{sec:forward-transfer}

We next examine how each model reuses knowledge learned from shorter tasks in these long-horizon settings.
\method{} reuses its complete pretrained predictor through the same language interface, whereas SARM replaces its task-specific classifier and SARM2 requires manual vocabulary mapping for full reuse.
For \method{}, we compare pretrained initialization with base Qwen3.5-4B while keeping the architecture, target data, adaptation procedure, and evaluation fixed.

\begin{table}[!htb]
  \centering
  \captionsetup{font=small,skip=5pt}
  \caption{Predictor reuse. Relative RMSE change (\%) from each method's baseline without reward-model pretraining; negative is better.}
  \label{tab:forward-transfer-deltas}
  \small
  \setlength{\tabcolsep}{4pt}
  \renewcommand{\arraystretch}{1.12}
  \begin{tabular*}{\linewidth}{@{\extracolsep{\fill}}llrrrrc@{}}
    \toprule
    & & \multicolumn{2}{c}{\textbf{Airplane}} & \multicolumn{2}{c}{\textbf{Car}} & \\
    \cmidrule(lr){3-4}\cmidrule(lr){5-6}
    \textbf{Method} & \textbf{Interface adaptation} & \textbf{D15}$\downarrow$ & \textbf{D100}$\downarrow$ & \textbf{D15}$\downarrow$ & \textbf{D100}$\downarrow$ & \textbf{Better} \\
    \midrule
    \multicolumn{7}{@{}l}{\textit{Full predictor reuse}} \\
    \addlinespace[2pt]
    \textbf{Ours} & \textbf{None required} & $-19.4$ & $-15.8$ & $-21.5$ & $-28.3$ & \textbf{4/4} \\
    SARM & Unsupported & --- & --- & --- & --- & N/A \\
    SARM2 & Manual mapping & $-9.7$ & $+4.4$ & $-18.2$ & $+2.2$ & 2/4 \\
    \midrule
    \multicolumn{7}{@{}l}{\textit{Partial predictor reuse}} \\
    \addlinespace[2pt]
    SARM & Adapt classifier & $+24.4$ & $-28.7$ & $-10.0$ & $-1.8$ & 3/4 \\
    SARM2 & Adapt classifier & $-2.9$ & $+22.2$ & $-9.1$ & $+15.6$ & 2/4 \\
    \bottomrule
  \end{tabular*}
  \par\smallskip
  \begin{minipage}{\linewidth}
    \footnotesize\raggedright
    D15 and D100 denote labeled target-trajectory budgets.
    \emph{Better} counts negative RMSE changes across the four task--budget settings.
    All methods receive subtask descriptions and target-task training.
    \emph{None required} means no additional task-interface changes.
    SARM's task-specific subtask classifier prevents full predictor reuse across different subtask sets; N/A and dashes indicate this architectural limitation.
  \end{minipage}
\end{table}

Pretraining reduces \method{}'s RMSE by 15.8--28.3\%, improving all four task--budget settings under full predictor reuse, compared with two or three for the tested baseline configurations (Table~\ref{tab:forward-transfer-deltas}).

\subsubsection{Trajectory Ranking on a Public Benchmark}
\label{sec:rbm-eval-ood}

We next test whether the learned progress model can rank complete trajectories on an external benchmark. Following the standard protocol in RBM-EVAL-OOD~\cite{liang2026robometer}, we report Kendall scores to measure agreement with the trajectory-quality ordering (failure $<$ suboptimal $<$ successful) via progress-only supervision fine-tuning.

Without reward-model pretraining, switching the backbone from Qwen3-VL to Qwen3.5-4B improves the macro ranking score only slightly, from 0.519 to 0.529 (Table~\ref{tab:rbm-eval-ood}).
Keeping Qwen3.5-4B fixed, pretraining with dense progress supervision on data from a different distribution brings a much larger gain, from 0.529 to 0.688, after the same progress-only fine-tuning on RBM-1M.
Both RoboMeter rows, including per-dataset values, are quoted from its paper.

\begin{table}[!htb]
  \centering
  \captionsetup{font=small,skip=5pt}
  \caption{RBM-EVAL-OOD ranking. Macro averages six domains;
  RoboMeter results quote arXiv v1~\cite{liang2026robometer}.}
  \label{tab:rbm-eval-ood}
  \label{tab:rbm-eval-ood-summary}
  \footnotesize
  \setlength{\tabcolsep}{3.5pt}
  \renewcommand{\arraystretch}{1.12}
  \begin{tabularx}{\linewidth}{@{}>{\raggedright\arraybackslash}Xrrrrrrr@{}}
    \toprule
    \textbf{Method / initialization}
      & \textbf{USC-F} & \textbf{USC-K} & \textbf{USC-T}
      & \textbf{USC-X} & \textbf{MIT-F} & \textbf{UTD-S}
      & \textbf{Macro}$\uparrow$ \\
    \midrule
    \multicolumn{8}{@{}l}{\textit{mixed-objective supervision}} \\
    RoboMeter: official \cite{liang2026robometer}
      & \textbf{0.646} & 0.471 & 0.653 & \textbf{0.694} & \textbf{0.601} & 0.867 & 0.655 \\
    \midrule
    \multicolumn{8}{@{}l}{\textit{progress-only supervision}} \\
    RoboMeter: progress-only \cite{liang2026robometer}
      & 0.083 & 0.231 & 0.333 & 0.389 & 0.183 & 0.533 & 0.292 \\
    Ours: no-pretrain (Qwen3-VL)
      & 0.333 & 0.456 & 0.389 & 0.500
      & 0.504 & \textbf{0.933} & 0.519 \\
    Ours: no-pretrain (Qwen3.5-4B)
      & 0.333 & 0.456 & 0.389 & 0.500
      & 0.560 & \textbf{0.933} & 0.529 \\
    Ours
      & 0.583 & \textbf{0.560} & \textbf{0.917} & 0.639
      & 0.496 & \textbf{0.933} & \textbf{0.688} \\
    \bottomrule
  \end{tabularx}
\end{table}

\FloatBarrier
\subsection{Subtask Composition and Progress Prediction}
\label{sec:pretraining-composition}

We test whether subtask composition improves progress prediction, then examine the roles of subtask selection and local-progress prediction.

\Needspace{175pt}
\subsubsection{Subtask Composition}
\label{sec:stage-selector-ablation}

\begingroup
\setlength{\intextsep}{3pt}
\setlength{\columnsep}{12pt}
\begin{wraptable}{r}{0.48\textwidth}
  \centering
  \captionsetup{font=footnotesize,skip=5pt}
  \caption{Subtask composition on airplane model assembly.}
  \label{tab:stage-selector-ablation}
  \footnotesize
  \setlength{\tabcolsep}{3pt}
  \renewcommand{\arraystretch}{1.12}
  \begin{tabular*}{\linewidth}{@{\extracolsep{\fill}}lrr@{}}
    \toprule
    \textbf{Configuration}
      & \textbf{D15}$\downarrow$
      & \textbf{D100}$\downarrow$ \\
    \midrule
    Direct whole-task
      & 0.462 & 0.359 \\
    Subtask composition
      & \textbf{0.382} & \textbf{0.311} \\
    \bottomrule
  \end{tabular*}
\end{wraptable}

On airplane model assembly, direct whole-task and composed predictors start from base Qwen without reward-model pretraining and use matched adaptation budgets.
Automatic subtask selection and local-progress composition reduce RMSE by 17.3\% with D15 and 13.4\% with D100, including selection errors (Table~\ref{tab:stage-selector-ablation}).

\FinishWrap
\endgroup

In zero-shot car model assembly, composition reduces accumulated RMSE from 1.578 to 0.524 for GPT-6 and by 32.4\% for base Qwen3.5-4B (Table~\ref{tab:hw-initialization-stage-progress}).
These gains show that our proposed subtask composition improves progress prediction in both models.

\paragraph{Inference latency and cost}
On HotWheels episodes, \method{}'s 4B model averages 155.95\,ms (approximately 0.16\,s) per scoring query on one NVIDIA H100 at batch size 1, with subtask selection and local-progress prediction executed sequentially.
GPT-6 composition takes 41.55\,s per query, making \method{} approximately $266\times$ faster per query in this setup (Appendix~\ref{app:inference-cost}).
Under OpenAI's standard API pricing, GPT-6 composition costs approximately US\$38.08 per trajectory at 1 frame per second for trajectories averaging 340 seconds.
\method{} runs locally, uses less than 10\,GB of GPU memory in BF16 at batch size 1, and incurs no API fees; local GPU costs are not included in this fee comparison.
These latency and resource requirements make repeated reward evaluation more practical for policy learning.

\Needspace{295pt}
\subsubsection{Subtask Selection and Local Progress}
\label{sec:selection-local-progress}

\begingroup
\setlength{\intextsep}{3pt}
\setlength{\columnsep}{12pt}
\begin{wraptable}{r}{0.52\textwidth}
  \centering
  \captionsetup{font=footnotesize,skip=5pt}
  \caption{Long-horizon progress prediction on car model assembly. Base Qwen is Qwen3.5-4B; zero-shot Ours uses reward-model pretraining.}
  \label{tab:hw-initialization-stage-progress}
  \label{tab:gpt6-composition}
  \footnotesize
  \setlength{\tabcolsep}{3pt}
  \renewcommand{\arraystretch}{1.12}
  \begin{tabular*}{\linewidth}{@{\extracolsep{\fill}}lrr@{}}
    \toprule
    \textbf{Model / procedure} & \textbf{Sel. acc.}$\uparrow$ & \textbf{RMSE}$\downarrow$ \\
    \midrule
    \multicolumn{3}{@{}l}{\textit{Zero-shot}} \\
    \addlinespace[2pt]
    GPT-6: direct & --- & 1.578 \\
    GPT-6: composition (ours) & \textbf{86.56\%} & 0.524 \\
    Base Qwen: direct & --- & 4.541 \\
    Base Qwen: composition (ours) & 25.37\% & 3.068 \\
    Ours: self-selected & 69.46\% & 1.380 \\
    Ours: GPT-selected & \textbf{86.56\%} & \textbf{0.494} \\
    RoboMeter: GPT-selected & \textbf{86.56\%} & 0.624 \\
    \midrule
    \multicolumn{3}{@{}l}{\textit{Task-related adaptation (15 trajs)}} \\
    \addlinespace[2pt]
    Ours: no-pretrain + adapt. & 93.33\% & 0.502 \\
    Ours: pretrained + adapt. & \textbf{94.70\%} & \textbf{0.394} \\
    \bottomrule
  \end{tabular*}
\end{wraptable}

Finally, we examine subtask selection and local-progress prediction in the zero-shot car model results.
Using its own subtask predictions without car model adaptation, pretrained \method{} reaches 69.46\% subtask selection accuracy and RMSE 1.3803, versus 25.37\% and 3.0676 for base Qwen3.5-4B with composition.
For the GPT-selected variant, pretrained \method{} uses the same subtask predictions as GPT-6 with composition.
Holding GPT-6's subtask predictions fixed, replacing its local-progress estimates with those from pretrained \method{} further reduces accumulated RMSE from 0.524 to 0.494, without car model adaptation (Table~\ref{tab:hw-initialization-stage-progress}).
This result highlights the value of \method{}'s learned within-subtask progress prediction under the same subtask selector.
This hybrid variant retains the GPT-6 subtask-selection API call; the latency and absence of API fees reported above apply to fully local \method{} inference.

\FinishWrap
\endgroup

\FloatBarrier
\subsection{Downstream Policy Improvement}
\label{sec:policy-improvement}

We test whether predicted progress improves policies through offline reinforcement learning and progress-weighted behavior cloning.
Appendix~\ref{app:mpc} additionally reports test-time model predictive control on tractor assembly, where \method{} decides when to plan and selects among imagined futures.

\Needspace{300pt}
\subsubsection{Zero-Shot Rewards for Offline Reinforcement Learning}
\label{sec:short-task-policy}

\begingroup
\setlength{\intextsep}{3pt}
\setlength{\columnsep}{12pt}
\begin{wrapfigure}{r}{0.49\textwidth}
  \centering
  \captionsetup{font=footnotesize,skip=3pt}
  \includegraphics[width=\linewidth]{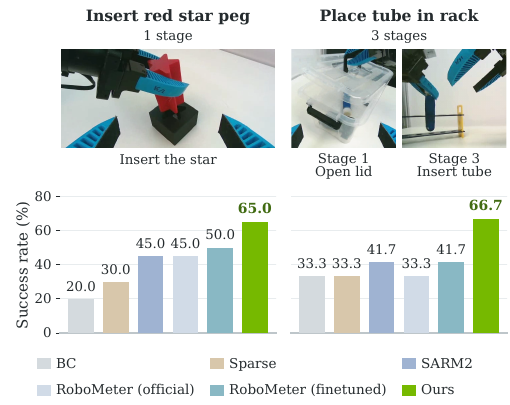}
  \caption{Policy success with zero-shot reward models on two held-out tasks.
  Each policy configuration is evaluated over 20 trials for star insertion and 12 for tube transfer.
  Across the four reward sources shared with Fig.~\ref{fig:scaling}(a), lower RMSE corresponds to higher or tied success on both tasks.}
  \label{fig:bppo-policy}
\end{wrapfigure}

Each reward source scores the same BC-policy rollouts on two held-out tasks: 50 for star insertion (one subtask) and 45 for tube transfer from a box to a rack (three subtasks).
We convert predicted progress into rewards using the same rule across all reward models.
These rewards supervise IQL critic learning~\cite{kostrikov2022offline}, followed by BPPO actor updates~\cite{zhuang2023behavior}.
Zero-shot means reward models receive no target-task reward annotations; policies train on scored target-task rollouts.

\method{} yields the highest observed success, 65.0\% on star insertion and 66.7\% on tube transfer, versus 50.0\% and 41.7\% for the strongest competing reward sources (Fig.~\ref{fig:bppo-policy}).
Across the four reward sources shared with Fig.~\ref{fig:scaling}(a), lower aggregate zero-shot RMSE on the 50 held-out tasks corresponds to higher or tied success on both downstream tasks.
This observed association supports using held-out progress RMSE to prioritize promising reward models for downstream policy evaluation.

\FinishWrap
\endgroup

\Needspace{225pt}
\subsubsection{Progress-Weighted Learning on Long-Horizon Tasks}
\label{sec:long-task-policy}

\begingroup
\setlength{\intextsep}{3pt}
\setlength{\columnsep}{12pt}
\begin{wrapfigure}{r}{0.49\textwidth}
  \centering
  \captionsetup{font=footnotesize,skip=3pt}
  \includegraphics[width=\linewidth]{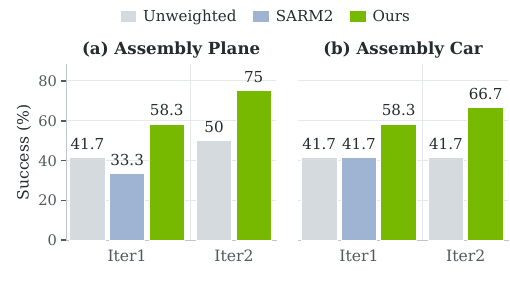}
  \caption{\textbf{Progress-weighted policy learning on long tasks.}
  Each reported success rate uses 12 evaluation trials.
  Comparisons share the task-specific parent policy within each iteration.
  }
  \label{fig:long-task-policy}
\end{wrapfigure}

On the airplane and car model assembly, adapted reward models weight recorded actions using the progress-difference procedure in Sec.~\ref{sec:policy-learning-method}.
We use a progress-delta reweighting scheme similar to SARM's reward-aligned BC~\cite{chen2025sarm}, keeping the weighting procedure fixed across reward-model comparisons.
We compare \method{} with unweighted learning and SARM2, the strongest adapted baseline by average RMSE on these two tasks at D100.

In Iter1, we test whether progress weighting improves learning from shared demonstrations and policy rollouts.
For each task, a BC policy trained on 20 demonstrations (25\% success) provides the common Iter1 initialization and collects 12 shared DAgger trajectories.
All conditions train on the original demonstrations plus these trajectories, using these progress-based weights or unweighted BC.
\method{} reaches 58\% success on both tasks, versus 42\% for unweighted BC (plain DAgger) and 33\%/42\% for SARM2 on airplane model and car model assembly, respectively.

In Iter2, we ask whether progress weighting still helps when all newly collected rollouts succeed.
We select the checkpoint with the highest observed Iter1 success rate for each task (58\%) as the shared parent for weighted and unweighted cloning.
Both conditions train on the original 20 demonstrations plus 12 newly collected successful rollouts from this parent, excluding the Iter1 DAgger trajectories; \method{} scores these rollouts for the same delta-progress weighting procedure.
Weighted cloning achieves 75\% versus 50\% on the airplane model task and 67\% versus 42\% on the car model task (Fig.~\ref{fig:long-task-policy}).
Since all 12 new rollouts succeed, terminal outcomes cannot distinguish useful from unhelpful chunks within these trajectories; the results support dense progress changes for local credit assignment.
\FinishWrap
\endgroup

\FloatBarrier
\section{Conclusion}
\label{sec:conclusion}

In this paper, we introduce \method{}, a generalizable reward model for robotics.
It combines structured progress supervision with a shared predictor conditioned on the subtask that should be the focus of evaluation.
Subtask boundaries and action-chunk judgments determine targets that capture failure resets and recovery through retries, and these targets also improve the existing RoboMeter architecture.
Pretraining on short tasks supports zero-shot reward prediction, long-task adaptation, and public OOD ranking after adaptation.
At the same time, the resulting progress estimates improve policies through offline RL and progress-weighted behavior cloning.
The observed agreement between rankings by held-out progress RMSE and policy success supports using RMSE to prioritize promising reward models for downstream evaluation.
The long-horizon comparisons show benefits from subtask-based inference in both GPT-6 and base Qwen3.5-4B, and from pretrained initialization of the 4B predictor under matched target-task adaptation.

\section*{Acknowledgments}
We used AI, including GPT-6 Astra, Claude Fable 5, and Opus 5, to polish the language, improve the presentation clarity of all figures, and assist with code implementation.

\bibliographystyle{unsrtnat}
\bibliography{references}

\begin{thebibliography}{29}
\providecommand{\natexlab}[1]{#1}
\providecommand{\url}[1]{\texttt{#1}}
\expandafter\ifx\csname urlstyle\endcsname\relax
  \providecommand{\doi}[1]{doi: #1}\else
  \providecommand{\doi}{doi: \begingroup \urlstyle{rm}\Url}\fi

\bibitem[Chen et~al.(2025)Chen, Yu, Schwager, Abbeel, Shentu, and
  Wu]{chen2025sarm}
Qianzhong Chen, Justin Yu, Mac Schwager, Pieter Abbeel, Fred Shentu, and
  Philipp Wu.
\newblock {SARM}: Stage-aware reward modeling for long horizon robot
  manipulation, 2025.

\bibitem[Liang et~al.(2026)Liang, Korkmaz, Zhang, Hwang, Anwar, Kaushik, Shah,
  Huang, Zettlemoyer, Fox, Xiang, Li, Bobu, Gupta, Tu, Biyik, and
  Zhang]{liang2026robometer}
Anthony Liang, Yigit Korkmaz, Jiahui Zhang, Minyoung Hwang, Abrar Anwar,
  Sidhant Kaushik, Aditya Shah, Alex~S. Huang, Luke Zettlemoyer, Dieter Fox,
  Yu~Xiang, Anqi Li, Andreea Bobu, Abhishek Gupta, Stephen Tu, Erdem Biyik, and
  Jesse Zhang.
\newblock Robometer: Scaling general-purpose robotic reward models via
  trajectory comparisons, 2026.

\bibitem[Alakuijala et~al.(2025)Alakuijala, McLean, Woungang, Farsad, Kaski,
  Marttinen, and Yuan]{alakuijala2025videolanguage}
Minttu Alakuijala, Reginald McLean, Isaac Woungang, Nariman Farsad, Samuel
  Kaski, Pekka Marttinen, and Kai Yuan.
\newblock {Video-Language Critic: Transferable Reward Functions for
  Language-Conditioned Robotics}.
\newblock \emph{Transactions on Machine Learning Research}, 2025.

\bibitem[Lee et~al.(2026)Lee, Wagenmaker, Pertsch, Liang, Levine, and
  Finn]{lee2026roboreward}
Tony Lee, Andrew Wagenmaker, Karl Pertsch, Percy Liang, Sergey Levine, and
  Chelsea Finn.
\newblock {RoboReward}: General-purpose vision-language reward models for
  robotics, 2026.

\bibitem[Chen et~al.(2026{\natexlab{a}})Chen, Harrison, Lee, Yang, Ren,
  Ratliff, Duan, Fox, and Krishna]{chen2026topreward}
Shirui Chen, Cole Harrison, Ying-Chun Lee, Angela~Jin Yang, Zhongzheng Ren,
  Lillian~J. Ratliff, Jiafei Duan, Dieter Fox, and Ranjay Krishna.
\newblock {TOPReward}: Token probabilities as hidden zero-shot rewards for
  robotics, 2026{\natexlab{a}}.

\bibitem[Chen et~al.(2026{\natexlab{b}})Chen, Zheng, Yu, Huang, Sun, Goldberg,
  Wen, Abbeel, Shentu, Wu, and Schwager]{chen2026sarm2}
Qianzhong Chen, Hau Zheng, Justin Yu, Suning Huang, Jiankai Sun, Ken Goldberg,
  Chuan Wen, Pieter Abbeel, Yide Shentu, Philipp Wu, and Mac Schwager.
\newblock {SARM2}: Multi-task stage aware reward modeling for self improving
  robotic manipulation, 2026{\natexlab{b}}.

\bibitem[Mao et~al.(2026)Mao, Yu, Mao, Li, Hu, Lan, Zhu, and Chen]{mao2026arm}
Yiming Mao, Zixi Yu, Weixin Mao, Yinhao Li, Qirui Hu, Zihan Lan, Minzhao Zhu,
  and Hua Chen.
\newblock {ARM}: Advantage reward modeling for long-horizon manipulation, 2026.

\bibitem[Ma et~al.(2023{\natexlab{a}})Ma, Sodhani, Jayaraman, Bastani, Kumar,
  and Zhang]{ma2023vip}
Yecheng~Jason Ma, Shagun Sodhani, Dinesh Jayaraman, Osbert Bastani, Vikash
  Kumar, and Amy Zhang.
\newblock {VIP: Towards Universal Visual Reward and Representation via
  Value-Implicit Pre-Training}.
\newblock In \emph{International Conference on Learning Representations},
  2023{\natexlab{a}}.

\bibitem[Ma et~al.(2023{\natexlab{b}})Ma, Liang, Som, Kumar, Zhang, Bastani,
  and Jayaraman]{ma2023livlanguageimagerepresentationsrewards}
Yecheng~Jason Ma, William Liang, Vaidehi Som, Vikash Kumar, Amy Zhang, Osbert
  Bastani, and Dinesh Jayaraman.
\newblock Liv: Language-image representations and rewards for robotic control,
  2023{\natexlab{b}}.

\bibitem[Rocamonde et~al.(2024)Rocamonde, Montesinos, Nava, Perez, and
  Lindner]{rocamonde2024visionlanguage}
Juan Rocamonde, Victoriano Montesinos, Elvis Nava, Ethan Perez, and David
  Lindner.
\newblock {Vision-Language Models are Zero-Shot Reward Models for Reinforcement
  Learning}.
\newblock In \emph{International Conference on Learning Representations}, 2024.

\bibitem[Sontakke et~al.(2023)Sontakke, Zhang, Arnold, Pertsch, B{\i}y{\i}k,
  Sadigh, Finn, and Itti]{sontakke2023roboclip}
Sumedh Sontakke, Jesse Zhang, S{\'e}b Arnold, Karl Pertsch, Erdem B{\i}y{\i}k,
  Dorsa Sadigh, Chelsea Finn, and Laurent Itti.
\newblock {RoboCLIP: One Demonstration is Enough to Learn Robot Policies}.
\newblock In \emph{Advances in Neural Information Processing Systems},
  volume~36, 2023.

\bibitem[Ma et~al.(2024)Ma, Hejna, Wahid, Fu, Shah, Liang, Xu, Kirmani, Xu,
  Driess, Xiao, Tompson, Bastani, Jayaraman, Yu, Zhang, Sadigh, and
  Xia]{ma2024generative}
Yecheng~Jason Ma, Joey Hejna, Ayzaan Wahid, Chuyuan Fu, Dhruv Shah, Jacky
  Liang, Zhuo Xu, Sean Kirmani, Peng Xu, Danny Driess, Ted Xiao, Jonathan
  Tompson, Osbert Bastani, Dinesh Jayaraman, Wenhao Yu, Tingnan Zhang, Dorsa
  Sadigh, and Fei Xia.
\newblock Vision language models are in-context value learners, 2024.

\bibitem[Huang et~al.(2026)Huang, Zhang, Hou, Huang, Su, Guo, Lu, Xu, Tang,
  Yang, Wang, Peng, Chen, Zhao, and Li]{huang2026rynnvalue}
Dongchi Huang, Hongyin Zhang, Bohan Hou, Siteng Huang, Zhian Su, Hang Guo, Tong
  Lu, Zhaofeng Xu, Jiahao Tang, Jianfei Yang, Donglin Wang, Peixi Peng, Mingxiu
  Chen, Deli Zhao, and Xin Li.
\newblock {RynnValue}: Scaling robotic value foundation models with temporal
  distance, 2026.

\bibitem[Zhang et~al.(2025{\natexlab{a}})Zhang, Luo, Anwar, Sontakke, Lim,
  Thomason, Biyik, and Zhang]{zhang2025rewind}
Jiahui Zhang, Yusen Luo, Abrar Anwar, Sumedh~Anand Sontakke, Joseph~J Lim,
  Jesse Thomason, Erdem Biyik, and Jesse Zhang.
\newblock {ReWiND}: Language-guided rewards teach robot policies without new
  demonstrations, 2025{\natexlab{a}}.

\bibitem[Yu et~al.(2026)Yu, Goldberg, Kondap, El-Refai, Ransing, Chen,
  Schwager, Shentu, Wu, and Goldberg]{yu2026warprm}
Justin Yu, Andrew Goldberg, Kavish Kondap, Karim El-Refai, Ethan Ransing,
  Qianzhong Chen, Mac Schwager, Fred Shentu, Philipp Wu, and Ken Goldberg.
\newblock {WARP-RM}: A warp-augmented relative progress reward model for data
  curation, 2026.

\bibitem[Liu et~al.(2026)Liu, Wen, Hu, Jayaraman, and Gao]{liu2026timerewarder}
Yuyang Liu, Chuan Wen, Yihang Hu, Dinesh Jayaraman, and Yang Gao.
\newblock {TimeRewarder}: Learning dense reward from passive videos via
  frame-wise temporal distance, 2026.
\newblock ICML 2026 spotlight; first preprint 2025.

\bibitem[Zhang et~al.(2026)Zhang, Zeng, Wang, Li, Lin, and
  Cheng]{zhang2026dream2reward}
Haoyu Zhang, Zecui Zeng, Bin Wang, Lusong Li, Liang Lin, and Long Cheng.
\newblock {Dream2Reward: Transition-Alignment Reward Models from Positive
  Demonstrations for Robotic Manipulation}, 2026.

\bibitem[Tan et~al.(2025)Tan, Chen, Xu, Wang, Ji, Chi, Lyu, Zhao, Chen, Co,
  Xie, Yao, Wang, Wang, and Zhang]{tan2025robodopamine}
Huajie Tan, Sixiang Chen, Yijie Xu, Zixiao Wang, Yuheng Ji, Cheng Chi, Yaoxu
  Lyu, Zhongxia Zhao, Xiansheng Chen, Peterson Co, Shaoxuan Xie, Guocai Yao,
  Pengwei Wang, Zhongyuan Wang, and Shanghang Zhang.
\newblock {Robo-Dopamine}: General process reward modeling for high-precision
  robotic manipulation, 2025.

\bibitem[Fang et~al.(2026)Fang, Dong, Liu, Yang, Huo, Mu, Yao, Li, Szafir, and
  Ding]{fang2026densereward}
Yu~Fang, Wanxi Dong, Jiaqi Liu, Yue Yang, Mingxiao Huo, Yao Mu, Huaxiu Yao,
  Li~Erran Li, Daniel Szafir, and Mingyu Ding.
\newblock {DenseReward}: Dense reward learning via failure synthesis for
  robotic manipulation, 2026.

\bibitem[Kim et~al.(2025)Kim, Heo, Lee, Lee, Shin, Lim, and Lee]{kim2025reds}
Changyeon Kim, Minho Heo, Doohyun Lee, Honglak Lee, Jinwoo Shin, Joseph~J. Lim,
  and Kimin Lee.
\newblock Subtask-aware visual reward learning from segmented demonstrations.
\newblock In \emph{International Conference on Learning Representations}, 2025.

\bibitem[Zhang et~al.(2025{\natexlab{b}})Zhang, Gao, Jiang, Chen, Pan, Zhang,
  and Deng]{zhang2025prm4rl}
Xiuhui Zhang, Ning Gao, Xingyu Jiang, Yihui Chen, Yuheng Pan, Mohan Zhang, and
  Yue Deng.
\newblock Progress reward model for reinforcement learning via large language
  models.
\newblock In \emph{Advances in Neural Information Processing Systems}, volume
  38, Main Conference, pages 100745--100772, 2025{\natexlab{b}}.
\newblock \doi{10.52202/085713-3370}.

\bibitem[Zhai et~al.(2025)Zhai, Zhang, Zhang, Huang, Zhang, Zhou, Zhang, Liu,
  Lin, and Pang]{zhai2025vlac}
Shaopeng Zhai, Qi~Zhang, Tianyi Zhang, Fuxian Huang, Haoran Zhang, Ming Zhou,
  Shengzhe Zhang, Litao Liu, Sixu Lin, and Jiangmiao Pang.
\newblock {A Vision-Language-Action-Critic Model for Robotic Real-World
  Reinforcement Learning}, 2025.

\bibitem[Kostrikov et~al.(2022)Kostrikov, Nair, and
  Levine]{kostrikov2022offline}
Ilya Kostrikov, Ashvin Nair, and Sergey Levine.
\newblock {Offline Reinforcement Learning with Implicit Q-Learning}.
\newblock In \emph{International Conference on Learning Representations}, 2022.

\bibitem[Zhuang et~al.(2023)Zhuang, Lei, Liu, Wang, and
  Guo]{zhuang2023behavior}
Zifeng Zhuang, Kun Lei, Jinxin Liu, Donglin Wang, and Yilang Guo.
\newblock {Behavior Proximal Policy Optimization}.
\newblock \emph{arXiv preprint arXiv:2302.11312}, 2023.

\bibitem[Peng et~al.(2019)Peng, Kumar, Zhang, and Levine]{peng2019advantage}
Xue~Bin Peng, Aviral Kumar, Grace Zhang, and Sergey Levine.
\newblock Advantage-weighted regression: Simple and scalable off-policy
  reinforcement learning, 2019.

\bibitem[{Qwen Team}(2026)]{qwen2026qwen35}
{Qwen Team}.
\newblock {Qwen3.5}: Towards native multimodal agents, February 2026.

\bibitem[Hu et~al.(2022)Hu, Shen, Wallis, Allen-Zhu, Li, Wang, Wang, and
  Chen]{hu2022lora}
Edward~J. Hu, Yelong Shen, Phillip Wallis, Zeyuan Allen-Zhu, Yuanzhi Li, Shean
  Wang, Lu~Wang, and Weizhu Chen.
\newblock {LoRA}: Low-rank adaptation of large language models.
\newblock In \emph{International Conference on Learning Representations}, 2022.

\bibitem[{OpenAI}(2026)]{openai2026gpt6}
{OpenAI}.
\newblock {GPT-6 Astra}.
\newblock OpenAI API documentation, 2026.
\newblock Accessed: 2026-09-16.

\bibitem[{NVIDIA et al.}(2026)]{nvidia2026gr00t2}
{NVIDIA et al.}
\newblock {GR00T 2}: Scaling robot foundation models for generalization,
  dexterity, and adaptation.
\newblock Whitepaper, NVIDIA, 2026.
\newblock URL
  \url{https://research.nvidia.com/labs/gear/gr00t-2/assets/gr00t2_whitepaper.pdf}.

\end{thebibliography}

\clearpage
\appendix
\section{Progress-Target Construction}
\label{app:target-construction}

This appendix expands the structured supervision in Sec.~\ref{sec:structured_supervision}.
We give the annotation inputs, the complete local-target construction, and the screw-insertion example underlying Fig.~\ref{fig:annotation-retry}.

\subsection{Annotation Inputs and Progress Composition}
\label{app:annotation-inputs}

Annotators provide subtask boundaries, action-quality (AQ) intervals, and a binary outcome for each subtask.
They mark intervals of suboptimal or regressed execution; unmarked intervals are nominal.
Nominal execution follows the task normally.
Suboptimal execution is degraded without physical failure, such as a misaligned screw that remains held.
Regressed execution loses local progress through physical failure, such as dropping the screw.
The outcome $o_k\in\{0,1\}$ records whether subtask $k$ ends in failure or successful completion.
A successfully completed subtask can contain regressed intervals followed by recovery.
These annotations do not require numerical progress scores for individual frames.

At global trajectory time $t$, let $k_t^\star$ be the zero-based annotated subtask index, $p_t\in[0,1]$ its local progress target, and $K$ the number of subtasks.
The whole-task target is
\begin{equation}
  P_t^\star=\frac{k_t^\star+p_t}{K}.
  \label{eq:appendix-target-composition}
\end{equation}
Each completed subtask contributes $1/K$.
At a completed boundary, the states $(k,1)$ and $(k+1,0)$ give the same whole-task progress.
A local reset therefore preserves the credit for preceding completed subtasks, under the assumption in Sec.~\ref{sec:stage_composition} that these subtasks remain completed.

\subsection{Chunk-Wise Local Progress}
\label{app:local-target-rule}

The construction uses three parameters (Table~\ref{tab:target-construction-parameters}).
The ratio $\rho$ controls how quickly nominal execution advances relative to suboptimal execution within an attempt.
The cap $p_{\mathrm{cap}}$ and reference duration $T_{\min}$ determine how far an attempt can advance before an observed regression resets it.

\begin{table}[!htb]
  \centering
  \caption{Parameters for local-progress construction. The cap and reference duration shown are for the screw-insertion subtask in Appendix~\ref{app:target-example}.}
  \label{tab:target-construction-parameters}
  \small
  \setlength{\tabcolsep}{5pt}
  \renewcommand{\arraystretch}{1.15}
  \begin{tabularx}{\linewidth}{@{}l>{\raggedright\arraybackslash}Xl@{}}
    \toprule
    \textbf{Symbol} & \textbf{Meaning} & \textbf{Value} \\
    \midrule
    $\rho>1$ & Nominal-to-suboptimal slope ratio & $20$ \\
    $0<p_{\mathrm{cap}}<1$ & Cap on the endpoint before regression & $0.95$ (screw insertion) \\
    $T_{\min}>0$ & Fastest measured time to reach the cap & $9\,\mathrm{s}$ (screw insertion) \\
    \bottomrule
  \end{tabularx}
\end{table}

\paragraph{Step 1: Locate and weight the chunk.}
Subtask boundaries and regressed AQ reset local progress.
Changes between nominal and suboptimal AQ do not reset it.
For an uninterrupted attempt, let $\tau_{\mathrm{start}}$ be its latest reset time and $\tau_{\mathrm{next}}$ the next regression onset or subtask boundary.
Both timestamps are measured in seconds from the start of the subtask.
The full chunk duration and time elapsed from its reset are
\begin{equation}
  T_{\mathrm{chunk}}=\tau_{\mathrm{next}}-\tau_{\mathrm{start}},
  \qquad
  t_{\mathrm{chunk}}=t_{\mathrm{subtask}}-\tau_{\mathrm{start}}.
  \label{eq:target-chunk-time}
\end{equation}
For each non-regressed update $i$ in the chunk $\mathcal C$, assign
\begin{equation}
  w_i=
  \begin{cases}
    \rho, & \text{nominal AQ},\\
    1, & \text{suboptimal AQ},
  \end{cases}
  \qquad
  N_{\mathrm{chunk}}=\sum_{i\in\mathcal C}w_i\,\Delta t_{\mathrm{chunk},i},
  \label{eq:target-weighted-duration}
\end{equation}
where $\Delta t_{\mathrm{chunk},i}=t_{\mathrm{chunk},i}-t_{\mathrm{chunk},i-1}$ and $t_{\mathrm{chunk},0}=0$ at the reset.
Thus, $N_{\mathrm{chunk}}$ is the weighted duration of the non-regressed updates.

\paragraph{Step 2: Set the endpoint and slope.}
Let $p_{\mathrm{last}}$ denote the target at the chunk's last non-regressed update.
The endpoint depends on how the chunk ends:
\begin{equation}
  p_{\mathrm{last}}=
  \begin{cases}
    p_{\mathrm{cap}}\min\!\left(1,\dfrac{T_{\mathrm{chunk}}}{T_{\min}}\right),
      & \text{Case A: interrupted by regression},\\[5pt]
    o_k, & \text{Case B: reaches the subtask boundary}.
  \end{cases}
  \label{eq:target-chunk-endpoint}
\end{equation}
In Case A, a chunk shorter than $T_{\min}$ reaches a proportionally smaller endpoint; a longer chunk reaches the cap.
In Case B, the annotated subtask outcome anchors the endpoint.
For either case, normalize the weighted updates using
\begin{equation}
  s_{\mathrm{chunk}}=\frac{p_{\mathrm{last}}}{N_{\mathrm{chunk}}}.
  \label{eq:target-chunk-slope}
\end{equation}

\paragraph{Step 3: Accumulate or reset.}
During a regressed interval, hold local progress at zero.
Otherwise, accumulate from the current chunk's reset:
\begin{equation}
  p_{t_{\mathrm{chunk},j}}
    =s_{\mathrm{chunk}}\sum_{i=1}^{j}w_i\,\Delta t_{\mathrm{chunk},i}.
  \label{eq:target-chunk-accumulation}
\end{equation}
Use this value as $p_t$ at the corresponding global time in Eq.~\ref{eq:appendix-target-composition}.
Nominal and suboptimal updates share the chunk's endpoint normalization, but their slopes differ by the factor $\rho$.
Recovery begins a new accumulation from zero while completed-subtask credit remains unchanged.

\subsection{Parameter Selection and a Worked Example}
\label{app:target-example}

For screw insertion, we compare successful demonstrations with deliberately induced failures across more than 20 diverse trials collected in advance.
Failures occur both while picking up the screw and just before insertion into the hole.
Comparing these events with successful demonstrations places the latest observed failures that reset local progress just below $0.95$ progress, motivating $p_{\mathrm{cap}}=0.95$ for this subtask.
The fastest observed time to reach this level is approximately $9\,\mathrm{s}$, giving $T_{\min}=9\,\mathrm{s}$.

\begin{figure}[!htb]
  \centering
  \includegraphics[width=\linewidth]{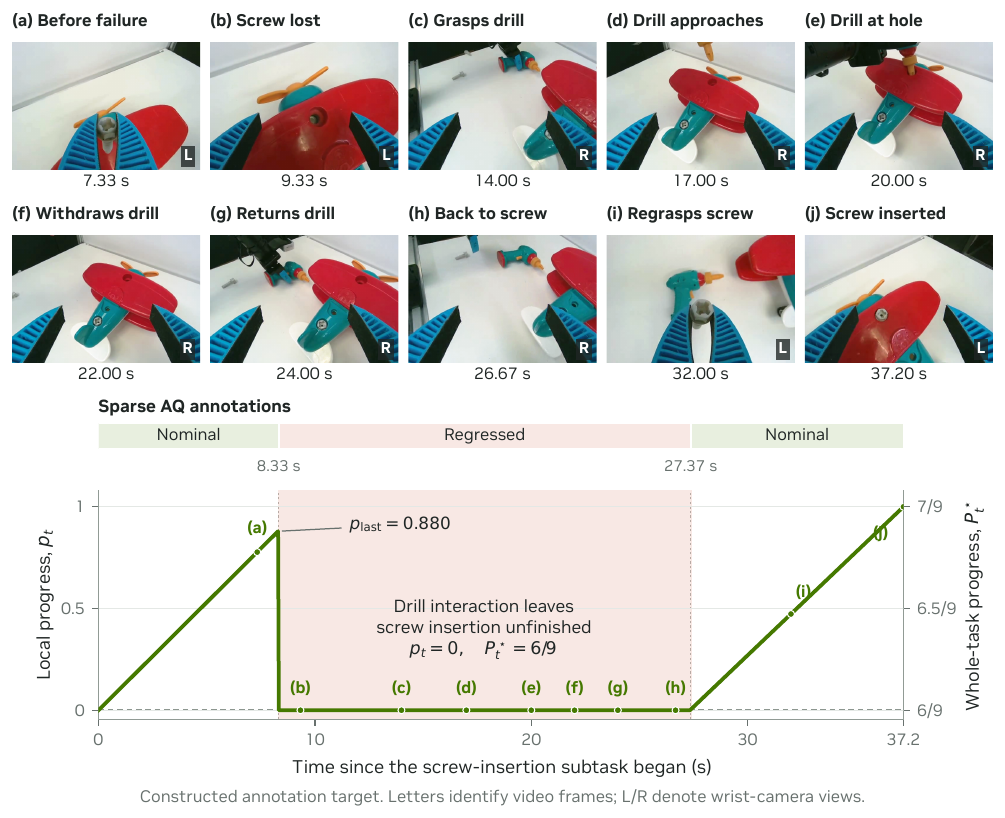}
  \caption{\textbf{Worked construction of a failure-and-recovery target.}
  Read the wrist-camera frames from left to right, then top to bottom; their letters mark the same times on the target curve.
  After losing the screw (b), the robot grasps the drill (c), brings it to the empty insertion hole (d--e), withdraws and returns it (f--g), and moves back toward the screw (h).
  These actions do not reinsert the screw, so local insertion progress remains zero and completed-subtask credit stays at $6/9$.
  The robot then regrasps the screw (i) and completes insertion (j), reaching $7/9$ whole-task progress.
  L/R denote the selected wrist-camera views.
  The curve is constructed supervision, not a model prediction; time is measured from the start of this subtask.}
  \label{fig:appendix-screw-target}
\end{figure}

Figure~\ref{fig:appendix-screw-target} follows insertion, subtask 7 of 9 in airplane assembly.
Six subtasks are already complete, so the whole-task target is $(6+p_t)/9$.
The example contains 1,117 synchronized samples at 30 frames per second, indexed from 0 through 1,116.
The first attempt is nominal through frame 249; frames 250--820 are regressed; frames 821--1,116 show nominal recovery to a successful outcome.

\paragraph{Before failure: Case A.}
Regression begins at $250/30\approx8.33\,\mathrm{s}$, so the interrupted chunk has $T_{\mathrm{chunk}}=250/30\,\mathrm{s}$.
Equation~\ref{eq:target-chunk-endpoint} gives
\begin{equation}
  p_{\mathrm{last}}=0.95\frac{250/30}{9}\approx0.8796.
  \label{eq:target-example-endpoint}
\end{equation}
The last non-regressed sample is frame 249, at $8.30\,\mathrm{s}$.
Normalizing over the non-regressed updates makes this sample reach $p_{\mathrm{last}}$ before the reset at frame 250.
Because every update in this attempt is nominal, the common factor $\rho$ cancels in Eq.~\ref{eq:target-chunk-accumulation}.

\paragraph{Regression and recovery: Case B.}
Frames 250--820 have $p_t=0$, retaining whole-task progress $6/9$.
During this interval, the robot picks up the drill and brings it to the empty insertion hole before returning it and reaching back toward the screw (Fig.~\ref{fig:appendix-screw-target}b--h).
Although the robot continues to move, these actions do not restore the unfinished screw-insertion subtask.
Recovery accumulates from the last zero-valued sample at $820/30\,\mathrm{s}$; its first positive update is frame 821.
The 296 nominal recovery updates reach the subtask boundary at frame 1,116, or $37.20\,\mathrm{s}$.
The successful outcome sets $p_{\mathrm{last}}=1$, and whole-task progress reaches $7/9$.

\subsection{Progress-Consistency Metrics}
\label{app:progress-consistency-metrics}

Table~\ref{tab:progress-consistency} scores the airplane and car model test episodes with whole-task progress (accumulated subtask progress divided by $K$, averaged over the four camera views) at about 1\,Hz.
Both metrics use the human annotations of these episodes.

\paragraph{Q-Rank.}
Annotators mark keyframes of high- and low-quality execution within each subtask.
For every pair of a high-quality and a low-quality keyframe of the same subtask from different episodes, we take the score at the evaluation point nearest to each keyframe (within 0.5\,s).
Q-Rank is the percentage of pairs in which the high-quality keyframe scores strictly higher.
Independent random scores drawn from the same continuous distribution rank either frame higher with equal probability, giving Q-Rank 50 on average.
The pairs involve 12 low-quality keyframes per task.

\paragraph{Down--Up (D--U).}
Each event is an annotated failure interval followed by a high-quality keyframe of the same subtask.
Let $s^-$ be the last score strictly before the failure onset.
We search the first 3\,s after that onset for the minimum score $s_{\min}$; this is the search window, not the duration of the annotated failure.
Let $s^+$ be the score nearest to the later high-quality keyframe (within 1\,s).
We count the score response to an annotated failure as correct when the window minimum is below both the pre-failure and recovery scores: $s^- > s_{\min}$ and $s^+ > s_{\min}$, with strict comparisons and no threshold.
Down--Up is the percentage of these annotated failure events with a correct score response.
Each event is counted once, even if the score drops repeatedly.

\paragraph{Reference values.}
A predictor whose scores never decrease cannot produce $s_{\min} < s^-$, so its Down--Up is 0.
A predictor whose score increases only with elapsed time is one such case.
Scores drawn independently from the same continuous distribution give Q-Rank 50 on average.
If $s^-$, $s^+$, and the $k$ window scores are distinct samples, the window minimum lies below both comparison scores with probability $k/(k+2)$, so random scores can satisfy the Down--Up criterion for many events.
A high Down--Up score alone therefore does not establish that score decreases are specific to failures.
We report Q-Rank alongside Down--Up to assess whether the scores also rank execution quality.

\FloatBarrier
\section{Reward Model and Training Details}
\label{app:model-training}

This appendix gives the implementation settings for the shared predictor in Secs.~\ref{sec:local_reward_model} and~\ref{sec:progress-training-details}.
Table~\ref{tab:reward-model-settings} summarizes the architecture, inputs, and optimization settings.

\subsection{Architecture and Inputs}

Subtask selection and local-progress prediction share the VLM, with a temporal transformer aggregating features for the progress head.
The sampled visual context can include later observations; the policy-learning experiments use offline scoring of recorded trajectories.

\begin{table}[!htb]
  \centering
  \caption{Reward-model implementation settings. The learning rate and batch size are shared across the compared reward-model training methods.}
  \label{tab:reward-model-settings}
  \small
  \setlength{\tabcolsep}{5pt}
  \renewcommand{\arraystretch}{1.18}
  \begin{tabularx}{\linewidth}{@{}p{0.27\linewidth}>{\raggedright\arraybackslash}X@{}}
    \toprule
    \textbf{Setting} & \textbf{Value} \\
    \midrule
    Backbone & Qwen3.5-4B vision--language model \\
    Frozen components & Base VLM weights and visual encoder \\
    Trained components & Language-decoder LoRA adapters, temporal transformer, and progress head \\
    LoRA rank & 32 \\
    Text inputs & Task instruction and ordered subtask descriptions \\
    Visual inputs & Eight sampled timesteps of synchronized camera views around the query time \\
    Calls per scoring query & Two: subtask selection, then progress prediction for the selected description \\
    Learning rate & $2\times10^{-5}$ \\
    Training batch size & 128 \\
    \bottomrule
  \end{tabularx}
\end{table}

\Needspace{110pt}
\subsection{Training and Inference}

Pretraining and adaptation use the objectives in Sec.~\ref{sec:progress-training-details}, retaining the same architecture and language interface.
During adaptation, descriptions selected by the model itself receive the order-conditioned targets in Eq.~\ref{eq:focus-progress-target}.

Inference uses two sequential calls: the first selects a subtask from the supplied ordered list, and the second predicts local progress conditioned on that subtask's description.
The selected index and local prediction are composed using Eq.~\ref{eq:composed-progress}.
Inference requires observations and text inputs, without subtask-boundary, AQ, or outcome annotations.

\subsection{Inference Latency and API Fees}
\label{app:inference-cost}

Table~\ref{tab:inference-latency-cost} reports latency per complete scoring query, including sequential subtask selection and local-progress prediction.
\method{}'s latencies are averages on HotWheels episodes, using one GPU at batch size 1.
GPT-6 latencies were measured over 8,000 API calls; GPT-6 denotes GPT-6 Astra-high as of September 15, 2026.
For GPT-6 composition, the per-query latency is the sum of the two stages: $29.00 + 12.55 = 41.55$\,s.

Each call receives eight sampled timesteps from four camera views, giving 32 image inputs per call and 64 across the two calls.
The 4B local model uses less than 10\,GB of GPU memory in BF16 at batch size 1 with this input configuration.
The reported times measure sequential query latency; processing independent queries in parallel can improve throughput.

\begin{table}[!htb]
  \centering
  \caption{Inference latency, speedup, GPU memory, and API fees. Speedup is $41.55\,\mathrm{s}/t_{\mathrm{query}}$, relative to a complete GPT-6 composition query. Local inference uses one GPU in BF16 at batch size 1. API fees are estimated at 1 frame per second for trajectories averaging 340\,s; local GPU costs are not priced. Dashes indicate unreported API-side memory or speedups not defined for component-only calls.}
  \label{tab:inference-latency-cost}
  \small
  \setlength{\tabcolsep}{4pt}
  \renewcommand{\arraystretch}{1.12}
  \begin{tabularx}{\linewidth}{@{}>{\raggedright\arraybackslash}Xrrrr}
    \toprule
    \textbf{Scoring procedure / hardware} & \textbf{Latency / query} & \textbf{Speedup} & \textbf{GPU VRAM} & \textbf{API fee / trajectory} \\
    \midrule
    \method{}: NVIDIA H100 & 155.95\,ms & $266.4\times$ & \cellcolor{nvidiagreen!12} & \cellcolor{nvidiagreen!12} \\
    \method{}: NVIDIA RTX 4090 & 451.06\,ms & $92.1\times$ & \cellcolor{nvidiagreen!12} & \cellcolor{nvidiagreen!12} \\
    \method{}: NVIDIA RTX 5090 & 334.19\,ms & $124.3\times$ & \cellcolor{nvidiagreen!12} & \cellcolor{nvidiagreen!12} \\
    \method{}: NVIDIA GB300 & 123.30\,ms & $337.0\times$ & \cellcolor{nvidiagreen!12}\multirow{-4}{*}{$<$10\,GB} & \cellcolor{nvidiagreen!12}\multirow{-4}{*}{No API fee} \\
    \midrule
    GPT-6 composition: total & 41.55\,s & $1.0\times$ & --- & $\approx$\,US\$38.08 \\
    \quad Subtask selection & 29.00\,s & --- & --- & $\approx$\,US\$20.31 \\
    \quad Local-progress prediction & 12.55\,s & --- & --- & $\approx$\,US\$17.77 \\
    \bottomrule
  \end{tabularx}
\end{table}

\FloatBarrier
\section{Offline Reinforcement Learning Details}
\label{app:offline-rl}

This appendix specifies the reward conversion and data used in the offline-RL experiments of Sec.~\ref{sec:short-task-policy}.
All reward sources score the same BC-policy rollouts within each task and use the same reward conversion.
The reward models receive no target-task reward annotations; policy learning uses the scored target-task rollouts.

\paragraph{Reward conversion and policy update.}
For each action chunk of horizon $H$, we sum discounted predicted whole-task progress increments:
\begin{equation}
  r_t(s_t,a_{t:t+H})=
    \sum_{j=0}^{H-1}\gamma^j
    \left[\hat P(s_{t+j+1})-\hat P(s_{t+j})\right],
  \qquad H=20,\quad\gamma=0.997.
  \label{eq:appendix-offline-reward}
\end{equation}
Here, $\gamma$ is the per-step discount factor.
These rewards supervise IQL action-value and state-value learning.
The resulting advantages $A(s,a)=Q_\phi(s,a)-V_\psi(s)$ guide BPPO actor updates, following Sec.~\ref{sec:policy-learning-method}.
Policy experiments use GR00T-2 as described in Sec.~\ref{sec:evaluation-protocol}, with initialization, data, and learning procedure held fixed within each reward-model comparison.

\begin{table}[!htb]
  \centering
  \caption{Offline-RL data and evaluation. Rollouts are shared across reward sources; evaluation counts are per policy configuration.}
  \label{tab:offline-rl-data}
  \small
  \setlength{\tabcolsep}{6pt}
  \renewcommand{\arraystretch}{1.15}
  \begin{tabularx}{\linewidth}{@{}>{\raggedright\arraybackslash}Xrrr@{}}
    \toprule
    \textbf{Task} & \textbf{Subtasks} & \textbf{Training rollouts} & \textbf{Evaluation trials} \\
    \midrule
    Star insertion & 1 & 50 & 20 \\
    Tube transfer from box to rack & 3 & 45 & 12 \\
    \bottomrule
  \end{tabularx}
\end{table}

\FloatBarrier
\section{Progress-Weighted Behavior Cloning Details}
\label{app:weighted-bc}

This appendix specifies the long-horizon policy-learning procedure in Sec.~\ref{sec:long-task-policy}.
We use the D100 versions of \method{} and SARM2, each fine-tuned with 100 labeled target-task trajectories.
Within each task and iteration, reward-model comparisons share the training data, parent policy, progress thresholds, and weighting rule.

\Needspace{160pt}
\subsection{Training Data and Parent Policies}

Table~\ref{tab:weighted-bc-data} summarizes the protocol for airplane model assembly ($K=9$) and red car model assembly ($K=10$).
In Iter1, all conditions start from the same BC parent and train on the demonstrations plus its shared DAgger trajectories.
In Iter2, the checkpoint with the highest observed Iter1 success for each task is the common parent for weighted and unweighted cloning.
The new rollouts all succeed and replace the Iter1 DAgger trajectories; the original demonstrations are retained.
Each reported policy success rate uses 12 evaluation trials per task.

\begin{table}[!htb]
  \centering
  \caption{Progress-weighted BC data and initialization, per task. Weighted and unweighted conditions share the parent and data within each iteration.}
  \label{tab:weighted-bc-data}
  \small
  \setlength{\tabcolsep}{5pt}
  \renewcommand{\arraystretch}{1.18}
  \begin{tabularx}{\linewidth}{@{}lp{0.22\linewidth}>{\raggedright\arraybackslash}Xr@{}}
    \toprule
    \textbf{Iteration} & \textbf{Parent policy} & \textbf{Training data} & \textbf{Eval. trials} \\
    \midrule
    Iter1 & BC on 20 demos & 20 demonstrations + 12 shared DAgger trajectories & 12 \\
    Iter2 & Best Iter1 checkpoint & Original 20 demonstrations + 12 new successful rollouts & 12 \\
    \bottomrule
  \end{tabularx}
\end{table}

\subsection{Progress Differences and Imitation Weights}

Let $\widehat P_t\in[0,1]$ be normalized whole-task progress at action time step $t$.
We compute
\begin{equation}
  \Delta\widehat P_t=\widehat P_{t+1}-\widehat P_t
  \label{eq:weighted-bc-increment}
\end{equation}
between adjacent frames at 30 frames per second, an interval of $1/30\,\mathrm{s}$.
Each action time step receives one imitation weight.
High, intermediate, and low progress increments receive weights $1$, $0.5$, and $0$, respectively, using the upper and lower thresholds in Table~\ref{tab:weighted-bc-thresholds}.
Actions assigned zero weight contribute no imitation loss.
For per-action imitation loss $\ell_t(\theta)$ and weights $\omega_t$, the weighted objective is normalized by the total weight:
\begin{equation}
  \mathcal L_{\mathrm{WBC}}(\theta)
    =\frac{\sum_t\omega_t\,\ell_t(\theta)}{\sum_t\omega_t},
  \qquad \omega_t\in\{0,0.5,1\}.
  \label{eq:weighted-bc-loss}
\end{equation}

\begin{table}[!htb]
  \centering
  \caption{Progress-weighted BC thresholds and weights. Threshold pairs are ordered as [upper, lower], in normalized whole-task progress units per action step.}
  \label{tab:weighted-bc-thresholds}
  \small
  \setlength{\tabcolsep}{5pt}
  \renewcommand{\arraystretch}{1.18}
  \begin{tabularx}{\linewidth}{@{}p{0.38\linewidth}>{\raggedright\arraybackslash}X@{}}
    \toprule
    \textbf{Setting} & \textbf{Value} \\
    \midrule
    Human-action thresholds & $[0.35,\,-0.13]\times0.01/K$ \\
    Policy-action thresholds & $[0.5,\,-0.2]\times0.01/K$ \\
    High / intermediate / low weights & $1.0\;/\;0.5\;/\;0$ \\
    Progress-difference interval & Adjacent frames at 30 fps ($1/30\,\mathrm{s}$) \\
    Subtask count $K$ & 9 (airplane); 10 (red car) \\
    \bottomrule
  \end{tabularx}
\end{table}

\FloatBarrier
\paragraph{Threshold selection.}
Thresholds are set from average nominal increments in progress curves constructed from human annotations in the reward-model evaluation data.
We then apply these thresholds to model-predicted increments on unseen policy-learning trajectories.
Separate human-action and policy-action thresholds reflect the greater diversity of initial conditions and motion strategies in human demonstrations.
The thresholds and weighting procedure remain fixed across the compared reward models.

\FloatBarrier
\section{Test-Time Model Predictive Control with \method{}}
\label{app:mpc}

Besides policy learning (Sec.~\ref{sec:policy-improvement}), predicted progress can guide a fixed policy at test time.
A world action model (WAM) predicts the outcome of an action chunk before executing it, so the robot can compare candidate actions before committing to one.
Model predictive control (MPC) with the WAM is costly, however: at each planning step the robot pauses while the WAM imagines candidate futures.
We therefore use \method{} in two roles at inference: it decides when to switch MPC on and off, and, while MPC is on, it selects among the imagined futures.
Both roles combine the selected subtask with its local progress, without additional training.

\subsection{Switching MPC On and Off}
\label{app:mpc-switch}

While the policy runs, \method{} scores the real camera stream every $\Delta t$ seconds and returns the selected subtask index $k_t$ and the composed whole-task progress $\hat P_t$ (Eq.~\ref{eq:composed-progress}).
A single misclassified frame should not trigger MPC, so we smooth the subtask index: the smoothed index $\bar k_t$ changes only after a new subtask is selected on $m$ consecutive estimates.
MPC switches on when either signal indicates that the policy is in trouble:
\begin{itemize}
  \item \textbf{Progress stall.} Whole-task progress has not grown over the last $W$ seconds, $\hat P_t-\hat P_{t-W}<\delta$: the policy is stuck, or repeated attempts are failing.
  \item \textbf{Subtask jump.} The smoothed subtask index skips ahead, $\bar k_t-\bar k_{t-\Delta t}>1$: the policy has moved to a later step before completing the current one, for example reaching for the drill before mounting a wheel.
\end{itemize}
MPC switches off once the robot has clearly moved past the difficult step: $\bar k_t$ advances to the next subtask and its local progress exceeds a margin $\epsilon$, or the final subtask's local progress approaches completion.
A cap of $M$ seconds of MPC keeps the robot from staying in planning mode when neither condition is met.
Control then returns to the policy alone until the next trigger.

\subsection{Selecting among Imagined Futures}
\label{app:mpc-select}

While MPC is on, the WAM samples $N$ candidate action chunks $a^{(n)}_{t:t+H}$ from different noise samples, each with a predicted video of its outcome.
If predicted progress trends downward for all $N$ candidates, we reject the batch and ask the WAM to generate a new batch.
On the final predicted frame of each candidate, \method{} selects the unfinished subtask and predicts its local progress, which Eq.~\ref{eq:composed-progress} composes into whole-task progress $\hat P^{(n)}_{t+H}$.
From an accepted batch, the robot executes the candidate with the highest $\hat P^{(n)}_{t+H}$ and replans from the new observation.

\subsection{Tractor Assembly}
\label{app:mpc-tractor}

\begin{figure}[!htb]
  \centering
  \includegraphics[width=0.42\linewidth]{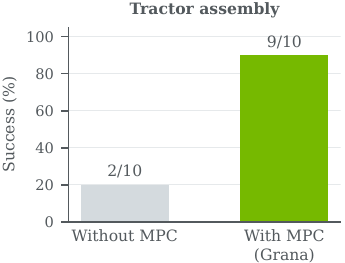}
  \caption{\textbf{Test-time MPC with \method{} on tractor assembly.}
  The same GR00T-2 policy runs from 10 different initial layouts with and without MPC.
  With MPC, \method{} switches MPC on and off and evaluates batches of four imagined futures.}
  \label{fig:tractor-mpc}
\end{figure}

We evaluate on tractor assembly, in which the robot mounts four wheels on a toy tractor body, fastens them with a drill, flips the body, and sets the tractor upright.
For fast inference, we use a \method{} variant with a Qwen3.5-0.8B backbone adapted to this task.
MPC is switched on and off by the rule in Sec.~\ref{app:mpc-switch}.
GR00T-2 (Sec.~\ref{sec:evaluation-protocol}) samples $N=4$ action chunks per round, each with a predicted 1.6\,s video; regeneration and selection follow Sec.~\ref{app:mpc-select}.

Without MPC, the policy completes 2 of the 10 assemblies; with MPC, it completes 9 of 10 (Fig.~\ref{fig:tractor-mpc}).
Failures without MPC include dropped or tilted wheels and skipped steps, such as reaching for the drill before mounting a wheel.
In the remaining MPC failure, all four wheels are mounted, but the tractor is not set upright.
These results indicate that \method{}'s subtask and progress estimates can both decide when planning is needed and guide action selection, without any policy update.

\end{document}